\pdfoutput=1

\documentclass[11pt, dvipsnames]{article}

\usepackage[final]{acl}

\usepackage{times}
\usepackage{latexsym}
\usepackage{amsmath}
\usepackage{dsfont}
\usepackage{booktabs}
\usepackage{multirow}
\usepackage{tcolorbox}
\usepackage{url}
\usepackage{hyperref}
\usepackage{caption}
\usepackage{subcaption}
\usepackage{bbm}
\usepackage{amssymb}

\usepackage[T1]{fontenc}

\newcommand{\ie}{i.e., }
\newcommand{\eg}{e.g., }

\usepackage[utf8]{inputenc}

\usepackage{microtype}

\usepackage{inconsolata}

\usepackage{graphicx}
\usepackage{pifont}

\usepackage{array}

\newcolumntype{H}{@{}>{\lrbox0}l<{\endlrbox}}
\title{Factual Knowledge in Language Models: Robustness and Anomalies under Simple Temporal Context Variations}

\author{
 \textbf{Hichem Ammar Khodja\textsuperscript{1,2}},
 \textbf{Frédéric Béchet\textsuperscript{2,3}},
 \textbf{Quentin Brabant\textsuperscript{1}},
\\
 \textbf{Alexis Nasr\textsuperscript{2}},
 \textbf{Gwénolé Lecorvé\textsuperscript{1}}
\\
 \textsuperscript{1}Orange - \textit{Lannion, France},\\
 \textsuperscript{2}Aix Marseille Université, CNRS, LIS, UMR 7020 - \textit{Marseille, France},\\
 \textsuperscript{3}International Laboratory on Learning Systems (ILLS - IRL2020 CNRS)
\\
 \small{
   \textbf{Correspondence:} \texttt{\{hichem.ammarkhodja, quentin.brabant, gwenole.lecorve\}@orange.com},
 }\\
 \small{
 \texttt{\{frederic.bechet, alexis.nasr\}@lis-lab.fr}
 }
}

\begin{document}
\maketitle
\begin{abstract}
This paper explores the robustness of language models (LMs) to variations in the temporal context within factual knowledge. It examines whether LMs can correctly associate a temporal context with a past fact valid over a defined period, by asking them to differentiate correct from incorrect contexts. The LMs' ability to distinguish is analyzed along two dimensions: the distance of the incorrect context from the validity period and the granularity of the context. To this end, a dataset called TimeStress is introduced, enabling the evaluation of 18 diverse LMs. Results reveal that the best LM achieves a perfect distinction for only 11\% of the studied facts, with errors, certainly rare, but critical that humans would not make. This work highlights the limitations of current LMs in temporal representation.
\end{abstract}

\section{Introduction}

When a Language Model (LM) completes the textual prompt "The capital of France is" with "Paris", it demonstrates that it has stored this fact somewhere in its parameters. However, as shown by numerous studies \cite{pararel,karr,robust_typo,robust_negation}, this type of factual knowledge is not necessarily robust to certain variations in the prompt (use of paraphrases, aliases, typographical errors, negations, etc.).
Among these variability factors, the temporal dimension of factual knowledge has been less studied. Thus, in this paper, we study the robustness of LMs' factual knowledge in the face of simple variations in the temporal context.

While the state of the art has demonstrated certain biases in LMs related to the temporal distribution of their training data or their weaknesses in reasoning with temporal concepts, our work aims to quantify how well LMs can correctly associate a temporal context (e.g., a year or a date, such as \textit{"In 2018, \dots"}, \textit{"On November 5, 2022, \dots"}) with a past fact, that is, a fact with a certain period of validity. More specifically, the research questions addressed are:
\begin{enumerate}
    \item Do LMs distinguish between correct and incorrect temporal contexts for facts?
    \item Do they differentiate them with the same accuracy depending on the distance of the incorrect context from the validity period of the facts?
    \item Do LMs activate their factual knowledge equally well when the temporal context is very precise or coarse?
\end{enumerate}

\begin{figure}[t!]
    \centering
    \includegraphics[width=1\linewidth]{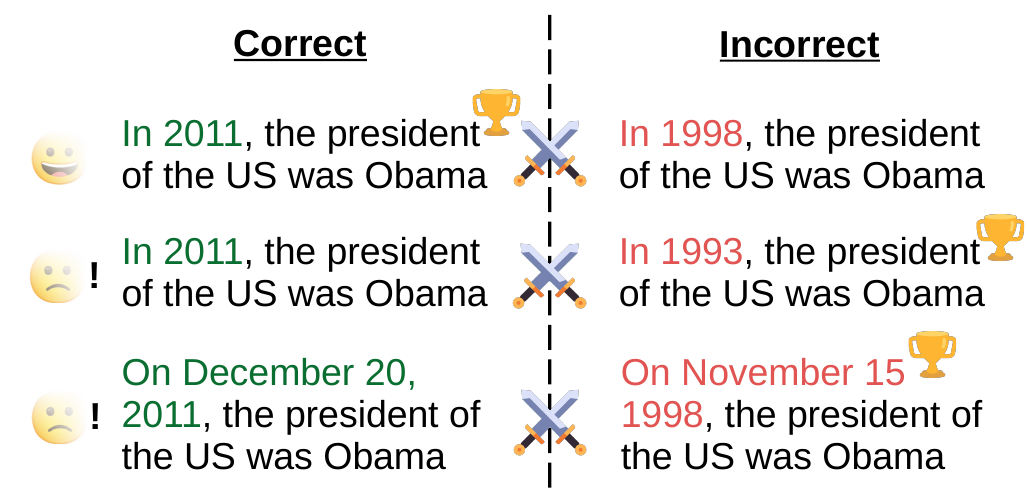}
    \caption{The robustness of the LM on a fact is evaluated by asking it to differentiate a set of correct and incorrect statements. The temporal context is varied along two dimensions: its position on the timeline (rows 1 and 2) and its granularity (rows 1 and 3). The \textit{trophy} means that the sentence was preferred by the LM.}
    \label{fig:robustness_illustration}
\end{figure}

To achieve this, as illustrated in Figure~\ref{fig:robustness_illustration}, matches are organized between correct and incorrect temporal contexts to measure the models' preferences, identify general trends, and highlight anomalies. As mentioned in the research questions, two specific angles of study are adopted to vary the temporal contexts within these matches: the positioning of the contexts on the timeline and their granularity (from the year to a specific date).

The contributions of the paper are:
\begin{itemize}
    \item The release of a dataset, \textit{TimeStress}, consisting of popular factual knowledge (according to a popularity index), temporally annotated, and their corresponding high-quality verbalizations. This dataset allows for the replication of our experiments but also opens avenues for other studies on temporality.
    \item Highlighting the low robustness of current LMs regarding their factual knowledge when it comes to positioning them in time, as well as errors—certainly rare but critical—that a human would not make. These results reveal the shortcomings of LMs in terms of internal representation of temporality, including for large models (18 models tested across various sizes and families).
\end{itemize}

In the following sections, we first discuss related work (Section \ref{sec:related_work}). Then, we elaborate on the paper's issues and present the TimeStress dataset (Section \ref{sec:problematique_et_donnees}). Finally, we describe our experiments and analyze their results (Section \ref{sec:experimentation}). The source code and data to reproduce our results will be published soon.
The source code enabling the reproduction of our experiments is published on GitHub\footnote{\href{https://github.com/Orange-OpenSource/TimeStress}{github.com/Orange-OpenSource/TimeStress} (MIT License)} and TimeStress is distributed in Hugging Face\footnote{\href{https://huggingface.co/datasets/Orange/TimeStress}{huggingface.co/datasets/Orange/TimeStress} (CC BY-SA 4.0 License)}.

\section{Related work}
\label{sec:related_work}

This section presents related work to ours, focusing on the study of factual knowledge in LMs, the consideration of their temporal aspect, and their temporal reasoning abilities.

\paragraph{Robustness of factual knowledge in LMs.} It has been demonstrated that LMs store a significant amount of factual knowledge \cite{petroni, know_what_language_models_know, head_to_tail}. However, numerous studies indicate that this acquired knowledge often lacks consistency when faced with textual perturbations. For example, \citet{robust_negation} highlighted the limitations of pretrained LMs in adapting to negations in questions, leading to contradictory answers. Robustness to paraphrasing and minor typographical errors has also been widely studied \cite{cc_paraphrase1, cc_paraphrase2, cc_paraphrase3, cc_paraphrase4}. Notably, \citet{pararel} and \citet{cc_paraphrase5} found that LMs produce different answers for semantically equivalent factual queries. Similarly, \citet{robust_typo} discovered that recent LMs can be negatively impacted by minor typographical errors that preserve the original semantics. 

\paragraph{Temporal alignment of knowledge in LMs.} Since factual knowledge is constantly evolving, studies have been conducted to understand how to adapt LMs to this evolution. As expected, LMs have been shown to be incapable of predicting future facts \cite{mindthegap}, highlighting the need to adapt them to maintain alignment with current knowledge. To address this issue, methods such as continual learning \cite{streaming_qa} and specific pretraining techniques have been proposed, including the joint modeling of text and its associated timestamp to facilitate the acquisition of new temporal knowledge \cite{templama}; knowledge editing techniques \cite{rome, grace, melo, zhang-etal-2023-large}; or simply externalizing knowledge into an external database accessible by the LM through retrieval-augmented generation \cite{rag}. In parallel, several datasets have been proposed to detect outdated facts in LMs \cite{outdated1_set_the_clock, outdated2_carpediem, outdated3_signal, outdated4_realtime, outdated5_dyknow}, and to update LMs' factual knowledge \cite{ammar-khodja-etal-2024-wikifactdiff, atoke, wikibigedit, chronoedit}.

\paragraph{Temporal reasoning in LMs.} Several studies have examined the temporal reasoning capabilities of LMs~\cite{zhang-choi-2021-situatedqa, chu-etal-2024-timebench,tempreason1_mentaqa,fatemi2024testtimebenchmarkevaluating, templama, xiong-etal-2024-large, su-etal-2024-living}. Notably, the works of \citet{tempreason_timequestions} and \citet{tempreason1_bench_temp_reason} each proposed a dataset in which LMs are invited to answer questions involving the understanding of the temporality of facts. While these studies share similarities with ours in terms of data (temporally annotated facts), their objectives and methodologies differ. These studies test the mastery of certain temporal logic operators (date calculations, comparisons, etc.) and evaluate the average performance of LMs based on a one-test-per-fact principle. In contrast, we focus not on reasoning ability but on the robustness of knowledge, that is, the ability of an LM to recall the same fact across various temporal contexts.

\section{Problem Statement and Dataset}
\label{sec:problematique_et_donnees}

The goal of this paper is to measure how robust a Language Model (LM) is to the temporal context associated with a fact. To achieve this, the proposed experimental protocol involves analyzing the LM's preferences when faced with correct or incorrect contexts for the same fact. This section first formalizes this problem and then presents the TimeStress dataset, which instantiates it.

\subsection{Problem Statement} \label{sec:problematique}

\paragraph{Facts and Temporal Contexts.}

Classically, we consider \textit{facts} as RDF triplets (subject, relation, object), denoted as $(s, r, o)$, where subjects and objects are entities or literals, and relations originate from an ontology~\cite{petroni, elsahar-etal-2018-rex}.
When dealing with \textit{temporal facts}, this representation is extended to include a validity period $[a, b]$, as done in other works~\cite{DBLP:conf/aaai/YinJY024, jain-etal-2020-temporal, tempreason1_bench_temp_reason}. For a quintuple $(s, r, o, a, b)$, the subject $s$ is connected to the object $o$ via the relation $r$ during the period from date $a$ to date $b$.
For example, (Barack Obama, president, USA, 20 January 2009, 20 January 2017) is a temporal fact.

We define the notion of a \textit{temporal context} as a time interval over which we wish to test the validity of a temporal fact. To reduce the number of possibilities and frame our work, we limit these time intervals to either \textit{entire years} (e.g., 1998, i.e., all days of the year 1998), an \textit{entire month} of a given year (\eg November 1998), or a \textit{specific date} (\eg November 15, 1998). Subsequently, these three distinct granularities will be denoted as Y for "Year," YM for "Year-Month," and YMD for "Year-Month-Day."

Considering a temporal fact~$f = (s, r, o, a, b)$, a temporal context~$\tau$ is said to be \textbf{correct} for~$f$ if $\tau$ is fully included in $[a, b]$ (i.e., $\tau \subseteq [a, b]$), \textbf{incorrect} if it is not included at all ($\tau \cap [a, b] = \varnothing$), or \textbf{transitional} otherwise ($\tau \cap [a, b] \neq \varnothing$ and $\tau \not\subseteq [a, b]$). For example, given the validity period $[2017, 2019]$, $2016$ is incorrect, $2017$ is transitional, and $2018$ is correct.

To assess the ability of an LM to distinguish a correct context $\tau^+$ from an incorrect context $\tau^-$ for a given temporal fact $(s, r, o, a, b)$, two textual statements are constructed respectively. The form of the statements adopted in our work is that of a question about the fact $(s, r, o)$ followed by its answer ("What is the $r$ of $s$? $o$") and prefixed by a verbalization of the temporal context $\tau^+$ or $\tau^-$.
For the example about Barack Obama, two possible contexts are $\tau^+ = 2011$ and $\tau^- = 1998$, producing the statements "\textit{In \textbf{2011}, who was the president of the USA\string? Barack Obama}" and "\textit{In \textbf{1998}, who was the president of the USA\string? Barack Obama}."

Finally, we say that an LM~$M$ distinguishes a correct context from an incorrect context when it assigns a higher probability to the answer $o$ given the statement with $\tau^+$ compared to conditioning on $\tau^-$, i.e., $\Pr_M(o|s,r,\tau^+) > \Pr_M(o|s,r,\tau^-)$. The details of the computation of $\Pr_M$ can be found in Appendix~\ref{app:prob}.

The overall estimation of this ability involves considering a large set of facts with varied entities, relations, and validity periods, and testing numerous pairs $(\tau^+, \tau^-)$ for each fact. To make the results of these matches interpretable, we impose that the contexts of the same pair have the same granularity (Y, YM, YMD).

\paragraph{Metrics.}

We introduce two metrics. Given a fact~$f$ and a model $M$, we express the results using a \textit{win rate} $\mathcal{W}(M, f) \in [0,1]$ of $M$ for $f$, which is the ratio of the number of times the model preferred a correct context over an incorrect context for the single fact $f$ to the number of tests performed. Additionally, a \textit{robustness} metric, denoted $\mathcal{R}(M, f)$, verifies that correct contexts consistently outperform incorrect ones, defined as: $\mathcal{R}(M, f) = \mathbbm{1}[\mathcal{W}(M,f) = 1]$ where $\mathbbm{1}[]$ is the indicator function. It is important to note that \textbf{transitional contexts are not used in any way for the calculation of these metrics}, as their validity is ambiguous. Given a set of facts, the average win rates and average robustness are denoted $\mathcal{V}(M)$ and $\mathcal{R}(M)$ respectively.

%
For segmentation purposes in the analyses, these global metrics can be restricted to tests conducted with temporal contexts of a specific granularity (Y, YM, or YMD).

Finally, to measure the distance of a context $\tau$ relative to the validity period $[a,b]$ of a fact, we calculate its \textit{relative position}, denoted $\alpha$, as the number of days between the midpoint of $[a,b]$ and the midpoint of $\tau$, divided by the number of days in $[a,b]$. Thus, $|\alpha| < \frac{1}{2}$ for correct contexts, and $|\alpha| > \frac{1}{2}$ for incorrect contexts. For transitional contexts, the value $|\alpha|$ is explicitly set to $\frac{1}{2}$.

\subsection{The TimeStress Dataset}
\label{sec:timestress_key}

We present the TimeStress dataset, which enables our study. This dataset contains over 521,000 statements (in the form of questions) generated from 2,003 temporal facts, covering 1,883 unique entities (1,385 unique subjects and 1,113 unique objects) and 86 relations. A brief sample is provided in Table~\ref{tab:verb_sample}.

On average, each fact is associated with $11$ correct temporal contexts and $74$ incorrect ones, distributed across the three granularities Y, YM, and YMD. \textbf{There are enough correct and incorrect contexts to make it nearly impossible for a random model to be robust on any fact by chance.}

In what follows, we briefly introduce how TimeStress was built, covering the quintuplet collection from Wikidata, their verbalization in natural language using GPT-4o, and how incorrect and correct contexts were sampled for each quintuplet in order to create statements.

A more detailed version of this section can be found in Appendix \ref{appendix:timestress}.

\subsubsection{Quintuplet Collection}


The quintuplet collection process begins with a preprocessed version of Wikidata provided in \citet{ammar-khodja-etal-2025-factual}.
This source also provides a measure of each entity's popularity, defined as the median number of human visits per month to the Wikipedia article associated with the entity in 2020. This measure is used to define the popularity index of a quintuplet, calculated as the geometric mean of the popularity of its object and subject. 
Although the popularity of the subject and object does not imply the popularity of the fact, this index remains an interesting tool for finding facts "known" by LMs, as it is shown empirically in the experiments.


We collect and filter Wikidata facts following this procedure: (1) All quintuplets with a validity period (\ie a start or end date mentioned) and whose objects are not literals, such as quantities and dates, are collected. (2) Quintuplets valid within two distinct periods are removed to simplify result analysis, as this allows all dates outside the validity period to be considered incorrect. (3) Quintuplets without a delimited validity period (\ie a start AND end date mentioned) are removed. (4) Only quintuplets that were \textbf{valid prior to 2021} are retained, as this ensures that all these quintuplets are past facts for all studied LMs. (5) Only the quintuplets that are valid for \textbf{longer than three years} are retained to ensure a minimal number of correct temporal contexts of Y granularity.
(6) We keep only the most popular quintuplets using the popularity index. This results in a set of 2,098 quintuplets with a varied set of 86~relations.

\subsubsection{Quintuplet Verbalization}

The process of generating statements from quintuplets is carried out using GPT-4o. 
First, a prompt instructs GPT-4o to generate four linguistically diverse questions from a given tuple (subject, relation, object, year), with the following guidelines: the question must be in the past tense, begin with ``In [YEAR],'', be stated in a simple and concise manner without any detail that could give clues about the answer. It should be directly followed by the answer, which is the object.
%
The quality of the generated questions was analyzed to identify and eliminate incorrect entries. Initially, out of the 2,098 facts intended for verbalization, 53 failed, and 64 questions mistakenly used the subject as the answer instead of the object. These erroneous cases were removed from the dataset, resulting in a total of 2,003 facts and $2003 \times 4 = 8012$ questions.
A random sample of 50 questions was manually evaluated to ensure the overall quality of the generated questions. The evaluation revealed that only 1 out of 50 questions was incorrect, while the remaining questions were perfectly constructed (Wilson confidence interval at 95\% = [0.85, 0.99]), which demonstrates the high quality of the questions in our dataset.
Finally, the temporal context was removed and \textbf{each fact is randomly assigned one of its four associated questions}. 

\subsubsection{Context Sampling}

For each fact, based on its validity interval $[a, b]$, centered on $m = \frac{a+b}{2}$ and of duration $d = b-a$\footnote{The median of dates (in day precision) is used to perform arithmetic operations between dates.}, temporal contexts at the Y granularity are uniformly sampled over the wider interval $[m - 5d, m + 5d]$ with a step of $0.05 \times d$. From these Y-granularity contexts, YM-granularity contexts are generated by randomly selecting a month. Similarly, YMD-granularity contexts are determined by choosing a random day from each YM-granularity context\footnote{This sampling does not produce erroneous dates such as February 29 for non-leap years, or April 31.}. This process creates a hierarchy among contexts derived from the same year for a given fact. Note that when a date $d_2$ is chosen from a higher-granularity date $d_1$, it is necessarily correct (or incorrect) if $d_1$ is. However, $d_1$ may be transitional while $d_2$ is correct or incorrect. In such cases, $d_2$ is excluded from the set of correct or incorrect dates. This \textbf{guarantees} that the number of correct and incorrect contexts does not vary by granularity, avoiding bias when comparing model robustness across granularities. The corresponding years of the produced contexts are mainly located in the contemporary period between 1800 and 2020 (Appendix \ref{appendix:additional_results}), because the popularity index used to select the facts in TimeStress draws more often recent facts.
To produce the statements of each fact that will be used to compute the metrics, its corresponding statement is prefixed with the previously sampled temporal contexts associated with the fact.

\begin{table*}[t!]
\small
    \centering
\resizebox{\linewidth}{!}{
\begin{tabular}{p{6.5cm}llHp{5.5cm}}
\toprule
\textbf{Temporal fact} & \textbf{Temp. Cont.} & \textbf{Status} & \textbf{Statement} & \textbf{Statement} \\
\midrule
(Betty Ford, spouse, Gerald Ford, 1948-10-15, 2006-12-26) & 1983-03-21 & Correct & On March 21, 1983, who was the spouse of Betty Ford\string? & On March 21, 1983, who was the spouse of Betty Ford\string? Gerald Ford \\
\midrule
(Beirut, country, Ottoman Empire, 1520, 1918) & 1759-05 & Correct & In May 1759, to which sovereign state did Beirut belong\string? & In May 1759, to which sovereign state did Beirut belong\string? Ottoman Empire \\
\midrule
(Jimmy Butler, member of sports team, Chicago Bulls, 2011, 2017-06-22) & 1989-06-17 & Incorrect & On June 17, 1989, which basketball team did Jimmy Butler belong to\string? & On June 17, 1989, which basketball team did Jimmy Butler belong to\string? Chicago Bulls \\
\midrule
(Samarkand, country, Soviet Union, 1922-12-30, 1991-08-31) & 1789-03-31 & Incorrect & On March 31, 1789, what was the sovereign state of Samarkand\string? & On March 31, 1789, what was the sovereign state of Samarkand\string? Soviet Union \\
\midrule
(United States of America, head of government, Andrew Johnson, 1865-04-15, 1869-03-04) & 1865 & Transitional & In 1865, who served as the head of government for the United States of America\string? & In 1865, who served as the head of government for the United States of America\string? Andrew Johnson \\
\midrule
(Chris Evans, unmarried partner, Minka Kelly, 2007-05, 2014-10) & 2014 & Transitional & In 2014, who was Chris Evans romantically involved with\string? & In 2014, who was Chris Evans romantically involved with\string? Minka Kelly \\
\bottomrule
\end{tabular}}
    \caption{
    Random sample of statements generated from various facts and temporal contexts in TimeStress.
    }
    \label{tab:verb_sample}
\end{table*}

\section{Experimentation}
\label{sec:experimentation}

This section details our experiments on the TimeStress dataset. As a reminder, our objectives are, in order, to measure the ability of models to distinguish correct and incorrect temporal contexts, analyze their robustness, and search for anomalies in this task when incorrect contexts are closer to or farther from the validity interval, and as the granularity of contexts becomes finer.

Numerous models from different families and sizes were tested: \texttt{Mistral-Nemo-Base-2407}, \texttt{Mistral-7B-v0.3} \cite{mistral}; \texttt{OpenEML-\{450M, 3B\}} \cite{openelm}; \texttt{gemma-2-\{2b, 9b, 27b\}} \cite{gemma2}; \texttt{Llama-3.1-\{8B, 70B\}} \cite{llama3}.
For each, both pretrained and instruction-tuned versions were considered, resulting in a total of 18 studied LMs.
All models were sourced from \textit{\href{https://huggingface.co}{huggingface.co}}.

In the first series of experiments, the statements were passed to the models as raw text rather than as instructions to enable the comparison between pretrained and instruction-tuned models. The use of an "instruction/message" format is explored in a second phase.

\subsection{Overall Mastery of Temporal Contexts}

\begin{figure*}[t]
    \centering
    \begin{subfigure}[h]{0.48\linewidth}
        \includegraphics[width=1\linewidth]{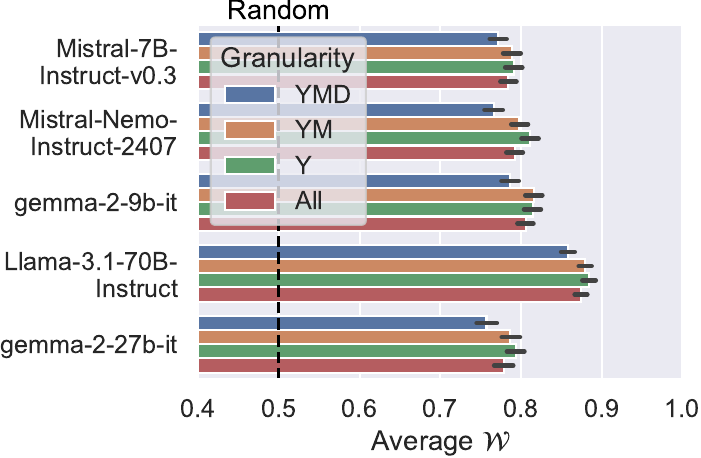}
        \caption{Average win rate}
        \label{fig:win_rate}
    \end{subfigure}\quad\begin{subfigure}[h]{0.48\linewidth}
        \vspace{2mm}
        \includegraphics[width=1\linewidth]{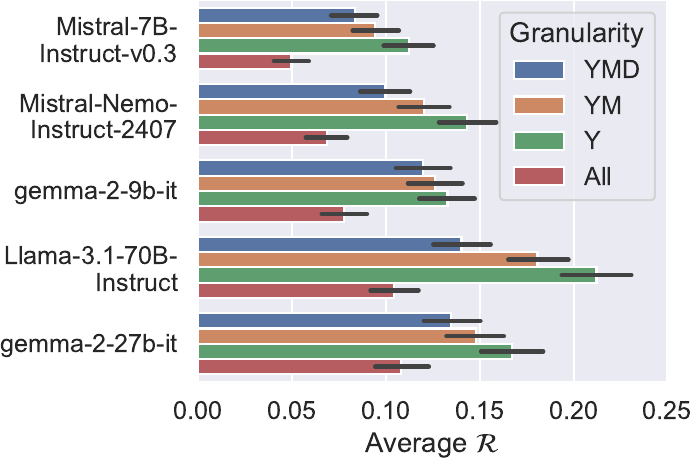}
        \caption{Average robustness}
        \label{fig:robust}
    \end{subfigure}
    \caption{Average metrics on the TimeStress dataset for the 5 most robust models (95\% CIs were determined using bootstrapping).}
\end{figure*}

Figure \ref{fig:win_rate} shows the average win rate for the facts in TimeStress for the top 5 LMs and for each temporal granularity Y, YM, and YMD, as well as for their union.
Results for other models are reported in Appendix~\ref{appendix:additional_results}.

Overall, the results show that these top 5 LMs generally distinguish correct statements from incorrect ones well, with win rates ranging from 78\% to 87\%. Among our other findings, we observed that even smaller models (<500M parameters) perform better than chance, and the win rate logically improves with model size (Appendix \ref{appendix:additional_results}), with the best model being the largest, Llama-3.1-70B-Instruct.

\begin{figure}[t]
    \includegraphics[width=\linewidth]{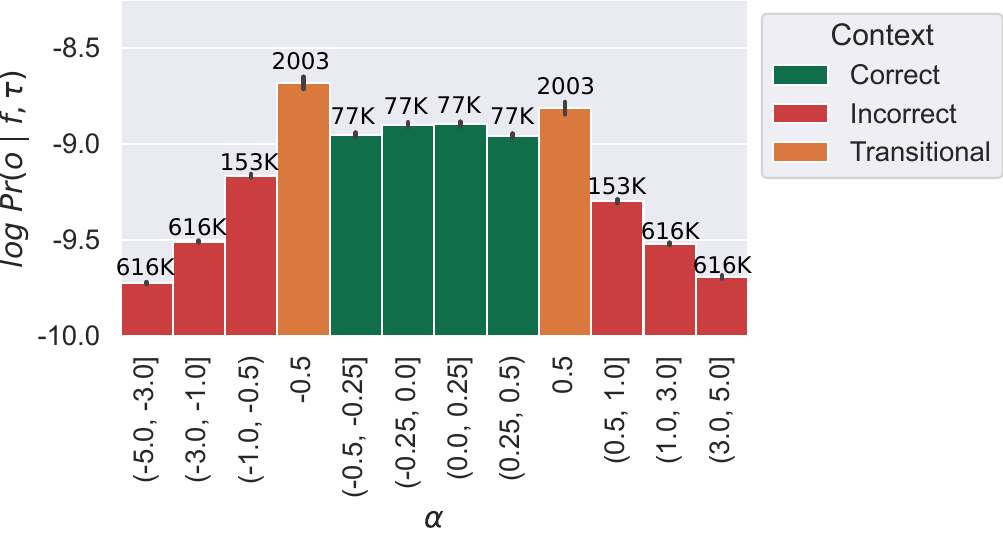}
    \caption{Evolution of $\log \Pr(o|f,\tau)$ with respect to the relative distance $\alpha$, averaged across all facts in TimeStress and all LMs, for granularity Y (Bootstrap 95\% CIs). The number of points used to compute each bar is indicated above it.}
    \label{fig:alpha_vs_logprob}
\end{figure}

Figure \ref{fig:alpha_vs_logprob} provides a more detailed analysis by reporting the average $\log \Pr(o | f, \tau)$ as a function of the value $\alpha$, which quantifies the relative distance of $\tau$ from the validity period of $f$ (see Section \ref{sec:problematique}).
The average is calculated across all facts, for contexts at the year granularity, and across all 18 studied LMs. We observe that the highest probabilities correspond to contexts within the validity interval ($\alpha \in [-0.5, 0.5]$), while outside this interval, probabilities gradually decrease as $|\alpha|$ increases.
Finally, we note that the probability assigned to transitional contexts (years that are neither fully correct nor fully incorrect) is significantly higher (based on the confidence intervals (CIs)) than that for incorrect contexts.
We explain this phenomenon with the following hypothesis: in the training data of LMs, transitional years are more often associated with the considered fact than other years within the validity period, as they correspond to key events such as the beginning and end of the fact (\eg the start or end year of a presidential term).

This strong alignment of LMs with the validity period of temporal facts leads us to conclude that LMs possess at least a basic representation of temporality.

\subsection{Robustness and Anomalies}
\label{sec:anomalies}

\paragraph{The temporal representation of LMs is not robust.}
Figure \ref{fig:robust} shows the average robustness of the top 5 models across all facts in TimeStress. As a reminder, this metric is stricter and does not tolerate any error during matches for a given fact. Results for other models are reported in Appendix~\ref{appendix:additional_results}. 

The scores are generally low, indicating that win rates per fact rarely reach 100\%. Interestingly, the most robust model is not the one with the highest win rate.
The most robust model, gemma-2-27b-it, achieves an $\mathcal{R}$ value of only about 17\% for the coarsest granularity Y. This score drops to 11\% when all granularities are considered. Most other models do not exceed a global robustness score of 3\%.
Among our other results, we also observed that instruction-tuned models mostly outperform their pre-trained counterparts. A notable case is the Llama-3.1-70B-Instruct model; although it was fine-tuned on instructions, it is $3.6 \times$ more robust than its pre-trained counterpart, Llama-3.1-70B. This suggests that the training data and possibly the training procedure play an important role in temporal robustness.
%
Finally, early signs of failure in knowledge transfer between granularities are evident due to the substantial gap between individual robustness scores for granularities and the global score. This issue is explored in detail later in this~section.

\begin{figure}[t]
    \centering
    \begin{tabular}{ccc}
    \toprule
    $|\alpha|$ & Raw Text & Instruction \\
    \midrule
    $\geq 1$ & 0.19 {\small(±0.01)} & 0.25 {\small(±0.01)} \\
    $\geq 2$ & 0.09 {\small(±0.01)} & 0.13 {\small(±0.01)} \\
    $\geq 3$ & 0.06 {\small(±0.01)} & 0.08 {\small(±0.01)} \\
    $\geq 4$ & 0.04 {\small(±0.01)} & 0.05 {\small(±0.01)} \\
    \bottomrule
    \end{tabular}
    \caption{Proportion of incorrect dates favored over correct dates beyond a relative distance $|\alpha|$, when the win rate exceeds 95\% (Wilson's 95\% CIs).}
    \label{tab:pos_incorrect_dates}
\end{figure}

\paragraph{LMs are vulnerable to \textit{easy} incorrect contexts.}
Table~\ref{tab:pos_incorrect_dates} investigates the impact of the relative positions of incorrect contexts of granularity Y, focusing on cases where incorrect contexts cause an LM to fail in a match for facts that seem "known" to the LM, as indicated by a very high win rate ($\mathcal{W} \geq$ 95\%). For now, only the "Raw Text" column is of interest. The table reveals that these incorrect contexts are not entirely concentrated around the validity period, as might reasonably be expected. Instead, a significant proportion is located far from it. 
Specifically, LMs fail to achieve robustness due to contexts with a distance of $|\alpha| \geq 1$ in 19\% of cases. This proportion decreases to 6\% for $|\alpha| \geq 3$, which remains significant given the proximity of the win rate to 100\% for the facts observed here. We conducted the same analysis using win rate thresholds higher than 95\% (see Appendix \ref{appendix:winrate_th}). As the threshold approaches 100\%, vulnerability to "easy" incorrect dates gradually decreases but never completely disappears. Even when the win rate threshold is 99\%, errors remain when $|\alpha| \geq 4$. We conclude that this vulnerability is inherent to current LMs. While the probabilistic nature of these models may provide a tangible explanation, this behavior is clearly undesirable, as these are typically errors that a human would not make when aware of a fact's validity period.

\begin{figure}[t!]
    \centering
    \includegraphics[scale=0.65]{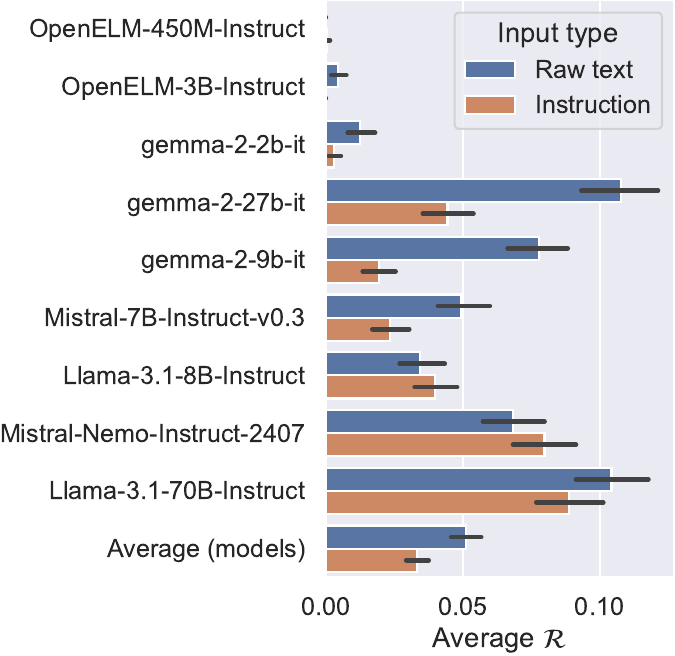}
    \caption{Average $\mathcal{R}$ across all granularities for facts in TimeStress based on the format of statements submitted to the models: raw text (blue) or instruction (orange). 95\% CIs were determined using bootstrapping.}
    \label{fig:inst_vs_noninst_inst_vs_noninst_query_robust}
\end{figure}

\paragraph{These conclusions hold for the instruction format.} 
So far in our experiments, all models have been fed statements in Raw text rather than instructions. Since the performance of instruction-tuned LMs might have been underestimated, win rates and robustness scores were recalculated using an "instruction/message" format\footnote{This involves constructing messages and injecting them into the chat template of each LM, as in the following example: \texttt{\{user: "In 2011, who was the president of the USA\string?", assistant: "Barack Obama"\}}.}. 
Figure \ref{fig:inst_vs_noninst_inst_vs_noninst_query_robust} compares robustness scores calculated for the two formats. On average, robustness decreases with the use of the "instruction" format (notably for gemma-2 models), and global robustness scores remain low. However, no clear conclusions emerge regarding the positive or negative impact of this format, as the effect varies significantly across models.
Next, the "Instruction" column of Table \ref{tab:pos_incorrect_dates} complements our previous analysis on the impact of the relative position of incorrect contexts for high win-rate facts. This time, the "instruction" format degrades performance with more critical errors (i.e., far from the validity period). Based on the confidence intervals, these differences are statistically significant for all values of $|\alpha|$ studied. Examples of these critical errors are shown in Appendix \ref{appendix:additional_results}.

\begin{figure}[t!]
    \centering
        \includegraphics[scale=0.35]{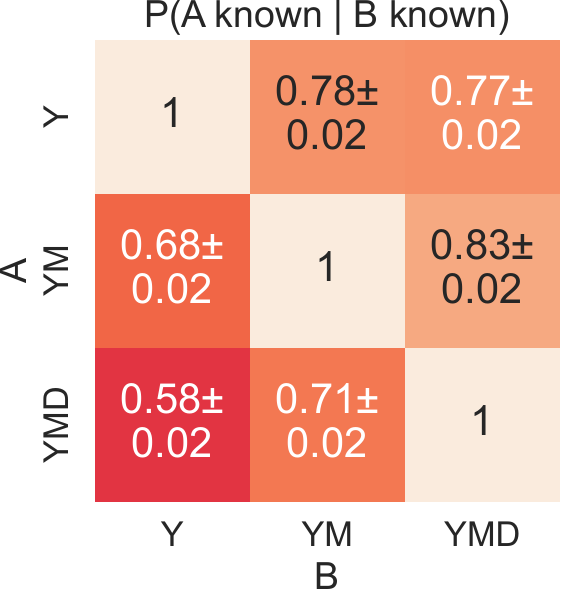}\ \includegraphics[scale=0.35]{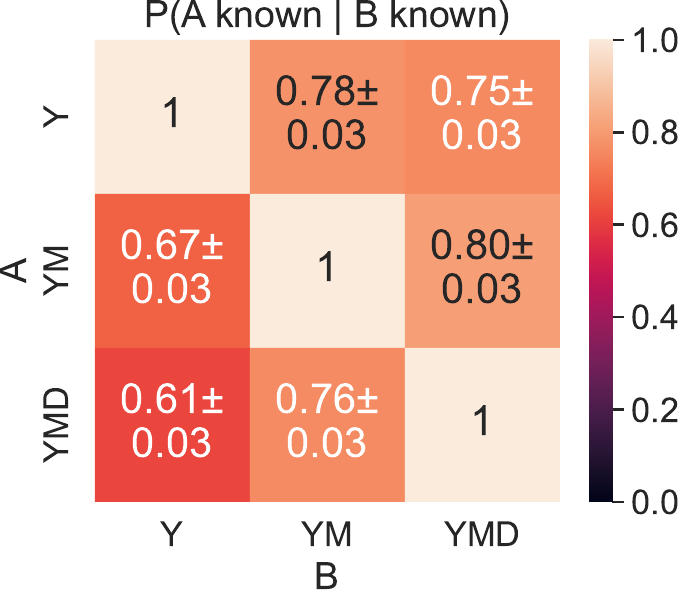}
        
        \caption{Average success rate of knowledge transfer between granularity pairs for the 5 most robust LMs with queries in \textbf{raw text (left)} or \textbf{instructions (right)}. Wilson confidence intervals at 95\% are shown.}
        \label{fig:gran_gen}
\end{figure}

\paragraph{LMs fail to perfectly propagate their knowledge across granularities.}
We examine the ability of LMs to propagate knowledge of a fact across different temporal granularities. TimeStress allows comparisons between two granularities because the three studied granularities have the same number of correct and incorrect contexts for all temporal facts. The only difference between two granularities is the addition of a random month and/or day, which does not affect validity when transitioning from a lower granularity to a higher granularity (\eg from Y to YM). For example, if a fact is incorrect for an entire year, it remains incorrect for any month or date within that year.

We consider a fact $f$ to be "known" for a granularity by a model $M$ if $\mathcal{R}(M,f)=1$. This definition can apply to a given granularity. For example, a fact is "known" at the Y granularity if all matches with temporal contexts at the year granularity were won. For each of the 5 most robust LMs and for each pair of granularities $(A,B)$, we then calculate the proportion of facts that are "known" at granularity $A$, given that they are "known" at granularity $B$.

Figure \ref{fig:gran_gen} reports this transfer proportion from granularity $B$ to $A$ for the "raw text" format (left) and "instruction" format (right). On average, for the "raw text" format, LMs failed to generalize their knowledge to other granularities in 28\% of cases (1 - average of all non-diagonal cells), which is surprisingly high given their perfect score on the starting granularity~$Y$. Details for each model are available in Appendix~\ref{appendix:generalization_matrices}. Performance varies across LMs. For example, for the most robust model, gemma-2-27b-it, the transition from $B=\mbox{Y}$ to $A=\mbox{YM}$ is successful in 74±5\% of cases, and the win rates for other transitions range between 68±6\% and 88±5\%. The general trend is that LMs fail more in transitions from coarse to fine granularities.
No LM achieves perfect transitions for any pair of granularities. 
There are slight variations between the instruction (Figure \ref{fig:gran_gen}, right) and raw formats, but the average success rate is nearly identical.

The possibility that poor knowledge propagation between granularities could be due to LMs' ignorance of the validity period boundaries\footnote{In this case, robustness was achieved only by chance.}. This was confirmed in a similar analysis that takes context position into account (Appendix \ref{app:corr}). Indeed, consistency between granularities approaches perfect consistency as the context moves away from the validity period. However, perfect consistency is \textbf{never reached}; which reminds us of the vulnerability of LMs to \textit{easy} incorrect contexts.

For exploratory purposes, we investigated whether including explanations about temporal concepts in the LMs' prompts could help them better transfer knowledge from one temporal granularity to another. To evaluate this, two prompts were prefixed to each TimeStress statement. The first explains the hierarchical nature of dates (\ie a year consists of months, and a month consists of days), while the second is more direct and explains how knowledge of a temporal fact can be generalized from one granularity to another. Details of these prompts are provided in Appendix~\ref{appendix:explanation_prompts}. We recalculated the transfer proportions between granularities using the same 5 LMs as in Figure~\ref{fig:gran_gen}. The two explanatory prompts improved generalization in the "raw text" format from 73\% to 76\%. However, no substantial gain compared to not using an explanatory prompt was observed when using the "instruction" format.

\paragraph{Other observations.} There is a positive correlation between the popularity of a fact and the robustness and win rate of LMs on it.
Interestingly, LMs are robust on globally different facts. Indeed, a pair of LMs shares, on average, 11\% of facts on which they are robust. This proportion reaches 31\% when limited to the 5 most robust LMs. However, only 34 facts out of 384 (8.9\%) are robust at the same time in these LMs.
Furthermore, the longer a fact's validity period, the higher the win rate (on the 5 most robust LMs). This statistically significant correlation\footnote{The null hypothesis is the absence of correlation.} is intriguing because it appears that the difficulty of situating a fact in time is the same whether it has a duration of 3 years or 30 years. 
One possible explanation is that facts with longer validity periods are more stable and unique (i.e., there are no alternative objects "o" for the same subject-relation pair "s,r"), so LMs can learn them without confusion or contradiction.
However, this explanation is contradicted by another observation: when there are more alternative objects "o" for a given (s,r) pair, the win rate and robustness actually increase, not decrease.
This contradiction raises the question of how to explain the observed phenomenon.
%
Finally, the further a fact's validity period is from the present, the less robust the LMs are on it, with lower win rates as well. More details are in Appendix \ref{appendix:additional_results}.

\section{Experimental Protocol: Motivations}
There are seemingly more "natural" approaches for probing factual knowledge in language models, such as the evaluation protocols used in LAMA \cite{petroni}, TriviaQA \cite{triviaqa}, KAMEL \cite{kamel}, and BEAR \cite{bear}. Instead of comparing probabilities across several temporal contexts, one could ask the LM to answer temporally contextual questions such as ``In 2011, who was the president of the US?'', and evaluate the LM based on the generated answers.
However, our experimental protocol
was preferred for several reasons.

First, our setup--where the LM must distinguish between statements with correct and incorrect temporal contexts by assigning probabilities--allows
to target specific facts without ambiguity, even in the case of
non-functional relations, such as "shares a border with," where a subject-relation pair can have multiple valid objects. In generation-based settings, an LM may produce one or several correct answers, or even off-topic outputs, making evaluation less reliable and direct comparison across LMs more difficult. This is especially true given that classical generation-based metrics, such as ROUGE \cite{rouge}, can underestimate performance. Sometimes, the set of all correct answers is difficult to enumerate due to the vagueness of the relation (\eg does asking for the borders of a country include continents and oceans?) and due to the sometimes large number of ways of expressing an answer.

Additionally, our evaluation protocol is efficient and scalable, as it does not require generation or answer validation.

Given the imperfections of other evaluation protocols, it would have been difficult to defend our claims--especially those involving sensitive metrics like robustness and the study of rare LM errors--if our results could be attributed to limitations of the evaluation method itself.

\section{Conclusion}
This study examined the robustness of LMs to simple temporal variations in factual knowledge. It assessed their ability to distinguish correct from incorrect temporal contexts based on two factors: the distance of contexts from the validity period of facts and their granularity. To facilitate this, the TimeStress dataset was introduced, featuring high-quality statements on popular temporal facts from Wikidata (according to a popularity index) and enabling the evaluation of 18 LMs of varying sizes and families.
The results revealed that the best-performing LM was robust for only 11\% of the studied facts, exhibiting errors, certainly rare, but critical that are uncommon to humans, which we frame as anomalies. These errors consist of a susceptibility to \textit{easy} incorrect contexts and imperfect knowledge generalization across granularities. Notably, these findings held true regardless of whether the LM was pretrained or instruction-tuned, and whether the statements were presented in an instruction or raw format. This highlights the limits of current LMs in temporal representation.
It is worth noting that since the studied temporal facts are relatively popular, these results likely represent an upper bound of LMs' performance on the general population of facts, given the strong link between knowledge popularity and its likelihood of being learned by LMs \cite{DBLP:conf/icml/KandpalDRWR23,kang-choi-2023-impact}.


\section*{Limitations}
The study evaluates LMs using a probability-based approach to assess their understanding of temporal facts. While this method does not fully capture model performance in text generation scenarios, it is strongly related, as generated text is sampled from the LM's probability distribution. Additionally, prior research has shown that probability-based metrics correlate reasonably well with the generative performance of models in factual knowledge evaluation contexts, where the model is expected to generate specific entities \cite{karr, lyu-etal-2024-beyond} as an answer, which is closely aligned with our experimental protocol.
The advantage of our approach compared to generation metrics is that it allows for precise exploration of specific non-functional relations where multiple  correct answers exist. This is more challenging with generation-based metrics, as LMs may produce another correct answer, unexpected responses, or off-topic outputs.

Second, the results of our study are limited to the format of the statements we chose, \ie a temporal context followed by a question and an answer. It is possible that LMs would perform better in a different format. However, their current limitations on our data are already problematic.

Finally, the TimeStress dataset consists of statements in English, which may limit the applicability of our results to other languages due to potential linguistic differences that could affect temporal understanding. However, future research can easily expand the scope by adapting the GPT-4o prompt used to generate statements to target additional languages. As for entity labels, they are available in other languages in Wikidata.

\bibliography{custom}

\appendix

\section{TimeStress: Details of the Construction Process}
\label{appendix:timestress}

This section provides a detailed description of the construction process for the TimeStress dataset. Before discussing the collection process, we describe the main characteristics of TimeStress.

First, the dataset focuses on past facts valid strictly before 2021, ensuring that they are historical (not valid at the present) events for all recent LMs. TimeStress includes high-quality statements that are consistent with the facts and exhibit linguistic diversity to avoid biases stemming from a limited variety of questions. The statements are carefully selected to minimize typographical errors, verbs are systematically conjugated in the past tense, and future dates beyond 2020 are excluded to avoid absurd questions such as "\textit{In 2052, who was the president of the USA\string?}". The dataset covers a diverse set of 86 relations to reduce biases associated with a restricted range. The targeted facts are popular, essential for evaluating the generalization of knowledge across different granularities—a task that becomes challenging if the LMs are unfamiliar with the facts. All facts are valid over a single validity period, ensuring that all contexts outside the validity period can be considered incorrect. Additionally, to ensure fairness, each granularity (Y, YM, YMD) has an equal number of correct and incorrect temporal contexts for all facts. Finally, the number of correct and incorrect contexts is sufficiently large to make it nearly impossible for a random model to be robust on any fact by chance.

The creation process for the TimeStress dataset was carefully designed to meet the properties described above, thereby effectively supporting the claims of this paper. This process consists of three main steps. First, an initial collection of 2,098 temporal facts is performed from Wikidata for inclusion in TimeStress. Second, questions are generated from these quintuplets using GPT-4o, accompanied by a quality evaluation to ensure high-quality questions. Finally, for each fact, correct and incorrect temporal contexts are identified and integrated into the questions to produce statements.

\subsection{Quintuplet Collection Process}
\label{app:quintuplet_collection}
The process of collecting quintuplets begins with the post-processed version of Wikidata provided by \cite{ammar-khodja-etal-2025-factual}.

This source also provides a measure of an entity's popularity, defined as the median number of human visits to the Wikipedia article associated with that entity during the year 2020. This measure is used to define the popularity of a quintuplet, calculated as the geometric mean of the popularity of its object and subject. Figure \ref{fig:pop_vs_robust} demonstrates the effectiveness of this popularity measure in identifying facts on which LMs are robust, illustrating that the likelihood of the robustness of LMs on a fact increases with its popularity.

Initially, all quintuplets with at least a start or end date and whose objects are not literals, such as quantities and dates, are collected, totaling over 2.1 million quintuplets. The quintuplets are then filtered to remove any $(s,r,o,a,b)$ where another quintuplet $(s,r,o,a',b')$ exists with a different validity period $[a',b']$, allowing us to assume that all dates outside $[a',b']$ are incorrect, which simplifies result analysis. This step eliminates a negligible amount of quintuplets (6.23\%). Additionally, quintuplets without a start or end date are removed as their validity period is unbounded.

Only quintuplets with a popularity measure of at least 90,000\footnote{This threshold was determined by gradually lowering the threshold from 150,000 in steps of 10,000 until the number of retrieved facts exceeded 2,000.} and a validity period strictly longer than three years are retained. 

The final result is a dataset comprising the 2,098 most popular facts from Wikidata (according to the popularity index), with 1,910 unique entities, 1,435 unique subjects, 1,151 unique objects, and 86 relations, forming a well-diversified set of temporal facts.

\subsection{Quintuplet Verbalization}
The process of verbalizing quintuplets into natural language questions is carried out using GPT-4o. The prompt, adapted from \citet{ammar-khodja-etal-2024-wikifactdiff} (Appendix B), was modified to generate questions instead of declarative sentences. The adapted \textit{system} prompt instructs GPT-4o to take a tuple (subject, relation, object, timestamp) and generate four linguistically diverse questions. For example, for the input \texttt{(British India, capital, Kolkata, 1929)}, a possible question could be: "\textit{In 1929, what was the capital of British India\string? Kolkata}". The questions must adhere to specific guidelines: they must be in the past tense, begin with the year followed by a comma, and end with the answer. The questions should focus on the object, be simple and concise, and avoid any detail that could simplify the answer.

\noindent Here is the \textit{system} prompt used:
\scriptsize
\begin{tcolorbox}[width=\columnwidth,colback=white]
\tt You are an advanced knowledge verbalization system.\\
You take as input a knowledge quadruple (subject, relation, object, time) and generate a list of 4 linguistically diverse questions on the quadruple.\\
For example, the input could be : (British India, capital, Kolkata, 1929) and one of your questions may be : "In 1929, what was the capital of British India\string? Kolkata.".\\

All the questions you generate must be in past tense because the facts are not valid anymore.\\
The questions must always start with the year, then a comma, then the question itself, and then finally the answer.\\
The questions must always be asked on the object.\\
The questions must be straightforward and concise.\\
The questions must not contain details that could make them easier to answer.\\

Examples of questions:\\
- (Jimmy Butler, member of sports team, Chicago Bulls, 2014) --> "In 2014, which team did Jimmy Butler play for\string? Chicago Bulls."\\
- (Philippines, head of state, Emilio Aguinaldo, 1900) --> "In 1900, who was the head of state of Philippines\string? Emilio Aguinaldo."\\
- (Coretta Scott King, spouse, Martin Luther King Jr., 1960) --> "In 1960, who was Coretta Scott King married to\string? Martin Luther King Jr."\\
- (European Union, currency, pound sterling, 2002) --> "In 2002, what was one of the currencies of the European Union\string? Pound sterling."\\
\end{tcolorbox}
\normalsize

\noindent And here is the main prompt:
\scriptsize
\begin{tcolorbox}[width=\columnwidth,colback=white]
\tt Here is the knowledge quadruple to verbalize: ([SUBJECT], [RELATION], [OBJECT], [YEAR]).\\

Due to the ambiguity that could arise from the provided labels, here is their meaning:\\
- (subject) "[SUBJECT]" : "[SUBJECT\_DESC]"\\
- (relation) "[RELATION]" : "[RELATION\_DESC]"\\
- (object) "[OBJECT]" : "[OBJECT\_DESC]"\\

Finally, here is an example where the relation "[RELATION]" is employed : ([EXAMPLE\_SUBJECT], [RELATION], [EXAMPLE\_OBJECT]).\\
\end{tcolorbox}
\normalsize

To use this main prompt, placeholders \texttt{[SUBJECT]}, \texttt{[RELATION]}, \texttt{[OBJECT]}, \texttt{[SUBJECT\_DESC]}, \texttt{[RELATION\_DESC]}, and \texttt{[OBJECT\_DESC]} are filled with the corresponding labels and descriptions from Wikidata. An example of the relation is also retrieved from Wikidata using the property \textit{Wikidata property example (P1855)}. If no example is available, the last line of the main prompt is omitted. The year [YEAR] is selected as the midpoint of the quintuplet's validity period. GPT-4o then generates four questions and answers for each quintuplet. Next, the temporal context is removed from the question, and it is verified that the answer matches the object.

\subsection{Quality of Generated Questions}

The quality of the generated questions was analyzed to identify and eliminate incorrect entries. Initially, out of the 2,098 facts intended for verbalization, 53 failed, and 64 questions mistakenly used the subject as the answer instead of the object. These erroneous cases were removed from the dataset, resulting in a total of 2,003 facts and $2003 \times 4 = 8012$ questions.

A random sample of 50 questions was manually evaluated to ensure the overall quality of the generated questions. The evaluation revealed that only 1 out of 50 questions was incorrect, while the remaining questions were perfectly constructed (Wilson confidence interval at 95\% = [0.85, 0.99])\footnote{This confidence interval was calculated with a finite population correction.}. These results demonstrate the high quality of the questions in our dataset.

Finally, each fact is randomly assigned one of its four associated questions.

\subsection{Test Generation}
Arithmetic operations between temporal contexts are involved in this section. It is important to note that all operations between contexts are performed on the midpoint of the context (as the contexts studied are intervals). For example, when $a + b$ is calculated, the result is the midpoint of $a$ added to the midpoint of $b$. The finest granularity a \textit{midpoint} can have is the YMD granularity (\ie Year-Month-Day). This approach bypasses the interval nature of dates.

For each quintuplet, the range of tested contexts is defined as $m \pm 5d$, where $m$ is the midpoint of the validity period $(a+b)/2$, and $d$ is the duration of the validity period $b-a$. To determine the dates of granularity Y (\ie Year) to include in TimeStress, we perform an analysis starting from the midpoint and extending to the boundaries with a step size of $0.05 \times d$. This step size is chosen to limit the maximum number of correct and incorrect contexts to reasonable values of 21 and 180, respectively.

For each context of granularity Y, a context of granularity YM is chosen by randomly selecting a month within the year. Similarly, for each context of granularity YM, a context of granularity YMD is chosen by randomly selecting a day within the previously selected YM context\footnote{This sampling does not produce erroneous dates such as February 29 for non-leap years, or April 31.}. This creates a hierarchical relationship between the different granularities (e.g., \textit{2020}, \textit{2020-03}, \textit{2020-03-24}), enabling reasonable comparisons in terms of win rates and robustness, as they share the same year and/or month. All contexts are now classified as correct, incorrect, or transitional (cf. Section \ref{sec:problematique}).

Despite this setup, a fact may have a variable number of correct and incorrect contexts per granularity due to transitional contexts, which may be absent in finer granularities if the $0.05 \times d$ step skips over them. This difference could bias performance, particularly favoring granularity Y in the robustness metric, which is calculated on fewer tests. To address this issue, YM-granularity and YMD-granularity contexts associated with transitional Y-granularity contexts are removed from the correct and incorrect sets and assigned to a special class called \textbf{Discarded}.

Finally, the contexts are converted into text and prefixed to the questions to create statements for each context at each granularity for each fact.

The resulting dataset, named \textbf{TimeStress}, includes 521,000 statements generated from 2,003 temporal facts. On average, it contains 11 correct dates and 74 incorrect dates, encompassing 1,883 unique entities, 1,385 unique subjects, 1,113 unique objects, and 86 relations. A random sample of TimeStress is presented in Table \ref{tab:timestress_sample}.

\section{Vulnerability to \textit{Easy} Incorrect Contexts: Analysis of Results at Different Win Rate Thresholds}
\label{appendix:winrate_th}

In Section \ref{sec:anomalies}, we demonstrated that LMs, even when they are almost robust on a fact (\ie a high win rate but inferior to 100\%), often fail to achieve robustness due to their vulnerability to easy contexts that are far outside the validity period (Table \ref{tab:pos_incorrect_dates}). 
In this section, we extend this analysis by experimenting with different win rate thresholds to observe how the distribution of incorrect contexts favored over correct contexts evolves as the threshold approaches 100\%.

\begin{figure*}[t]
    \centering
    \begin{subfigure}[h]{0.4\linewidth}
        \centering
        \includegraphics[width=\linewidth]{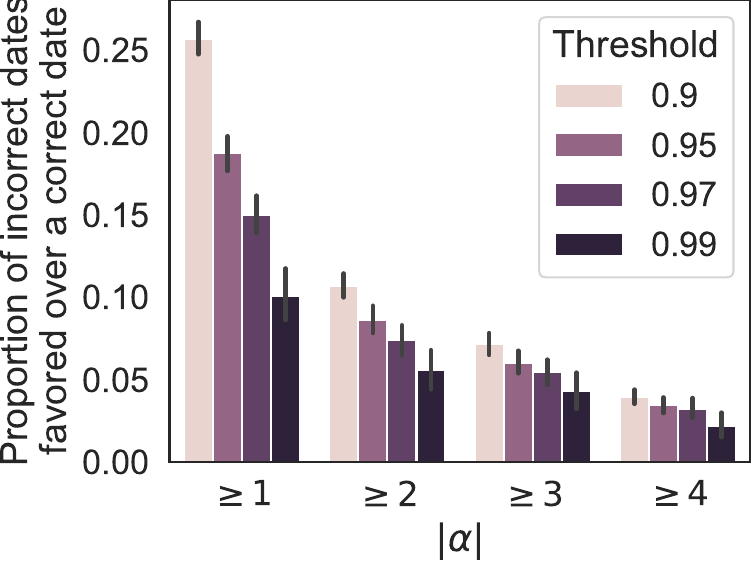}
        \caption{Raw Text}
    \end{subfigure}
    \ \ \ \ \ \ \ 
    \begin{subfigure}[h]{0.4\linewidth}
        \centering
        \includegraphics[width=\linewidth]{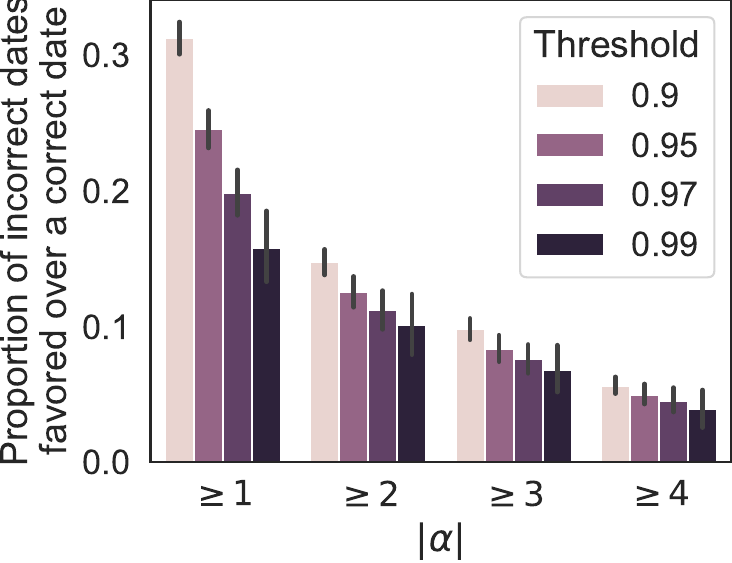}
        \caption{Instruction Format}
    \end{subfigure}
    \caption{Proportion of incorrect contexts favored over correct contexts that are beyond a relative distance $\alpha$ from the validity period, when the win rate exceeds the threshold, for the 5 most robust LMs. Experiments were conducted with granularity Y. 95\% confidence intervals were calculated using \textit{bootstrapping}.}
    \label{fig:appendix_th_alpha}
\end{figure*}

The results in Figure \ref{fig:appendix_th_alpha} indicate that even as the threshold approaches 1, LMs remain vulnerable to \textit{easy} incorrect contexts that are significantly distant from the validity period. We would expect LMs to definitively exclude highly distant contexts once they have acquired sufficient information about the validity period. However, this is not the case here, as even when the win rate is very close to 1, LMs continue to fail on these contexts. These results suggest that language models may never achieve true robustness, as the proportion of incorrect contexts converges toward zero but never fully reaches it. This implies that there will always be a possibility for an LM to fail on a distant incorrect context. This last point suggests that the already low percentage of robust facts could be even lower if we increased the number of incorrect and correct contexts used to calculate robustness.

\section{Generalization of Knowledge Across Granularities}
\label{app:granularites}

This section provides additional details and results regarding the generalization of knowledge across granularities.

\subsection{Consistency Across Granularities Based on Relative Distance}
\label{app:corr}

In this section, we examine the consistency of LM predictions across different granularities (Y, YM, YMD) as the distance between the tested context and the validity period increases.

To evaluate this, and solely for this section, we introduce a metric called local robustness. Local robustness for a fact, a LM, and a given incorrect context is equal to 1 if all correct contexts are preferred over this incorrect context, and 0 otherwise.

\begin{figure*}[t]
    \centering
    \includegraphics[width=0.82\linewidth]{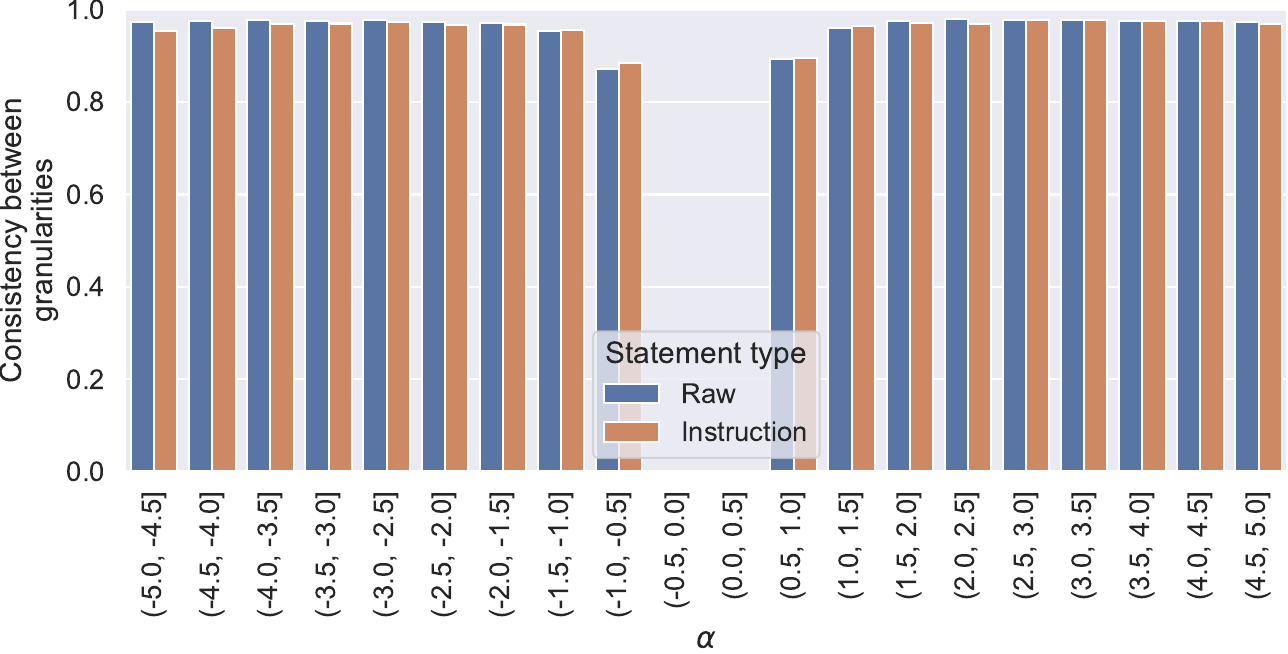}
    \caption{For each $\alpha$ segment, the average local robustness correlation across all granularity pairs is calculated over all facts and the 5 most robust LMs.}
    \label{fig:granularity_consistency}
\end{figure*}

We group all statements in TimeStress according to the relative distance $\alpha$ from their temporal context, and restricting ourselves to the 5 most robust MLs and to the "known" facts\footnote{We recall that "known" in the context of this article means that the ML in question has a robustness equal to 1 on the fact in question, \ie all correct contexts are preferred to incorrect contexts by the ML.} at least on one granularity by these LMs. These statements are categorized according to the interval of which their relative distance $\alpha$ is part. The chosen intervals are $]s, s+\frac{1}{2}]$, where $s$ can take values from $\{-5, -4.5, \dots, 4.5\}$. For each interval, the contexts are aligned by fact and by granularity hierarchically (\eg \texttt{2020}, \texttt{2020-04}, \texttt{2020-04-23}), which is guaranteed to be possible due to the properties of TimeStress (cf. Section \ref{sec:timestress_key}). Local robustness is then calculated for each incorrect context, and the accuracy\footnote{Accuracy measures the proportion of identical elements between two vectors, that is, the number of positions where the values are equal, divided by the total number of elements compared.} between these measures is computed for all granularity pairs (\ie Y-YM, Y-YMD, and YM-YMD). These coefficients are averaged across all granularity pairs, all facts, and the 5 most robust LMs, with the results presented in Figure \ref{fig:granularity_consistency}.

The results indicate that the inconsistency between granularities is mainly caused by incorrect contexts located at the boundaries of the validity period. As the context moves away from the validity period, the consistency approaches a perfect score of 1 but never reaches it regardless of the ML, the statement type and the $\alpha$ interval used.

\subsection{Generalization Matrices for Each LM}
\label{appendix:generalization_matrices}

\begin{figure*}
    \centering
    \begin{subfigure}[h]{0.195\linewidth}
        \centering
        \includegraphics[width=\linewidth]{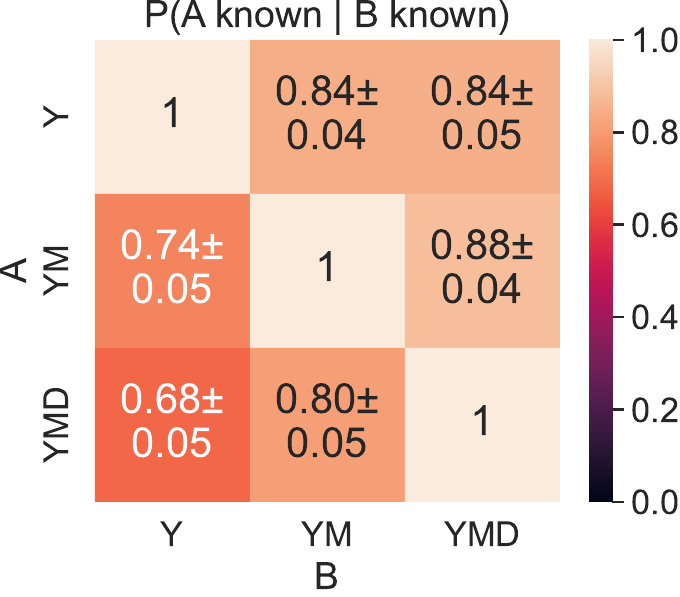}
    \end{subfigure}
    \begin{subfigure}[h]{0.195\linewidth}
        \centering
        \includegraphics[width=\linewidth]{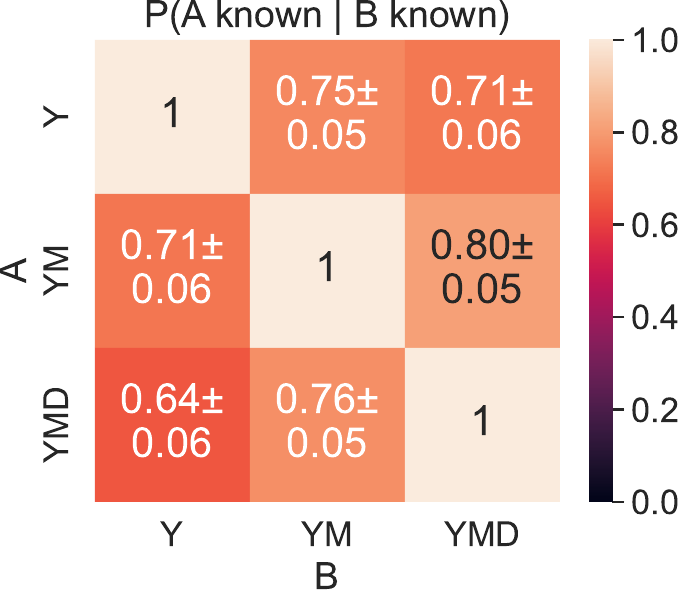}
    \end{subfigure}
    \begin{subfigure}[h]{0.195\linewidth}
        \centering
        \includegraphics[width=\linewidth]{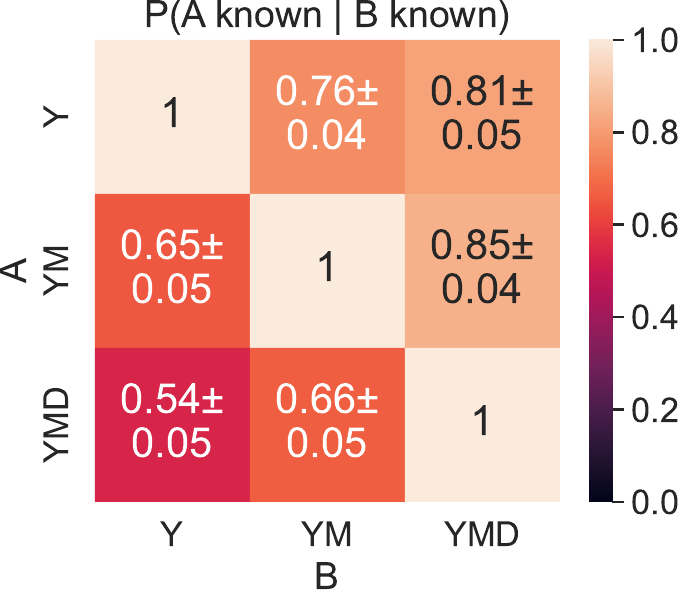}
    \end{subfigure}
    \begin{subfigure}[h]{0.195\linewidth}
        \centering
        \includegraphics[width=\linewidth]{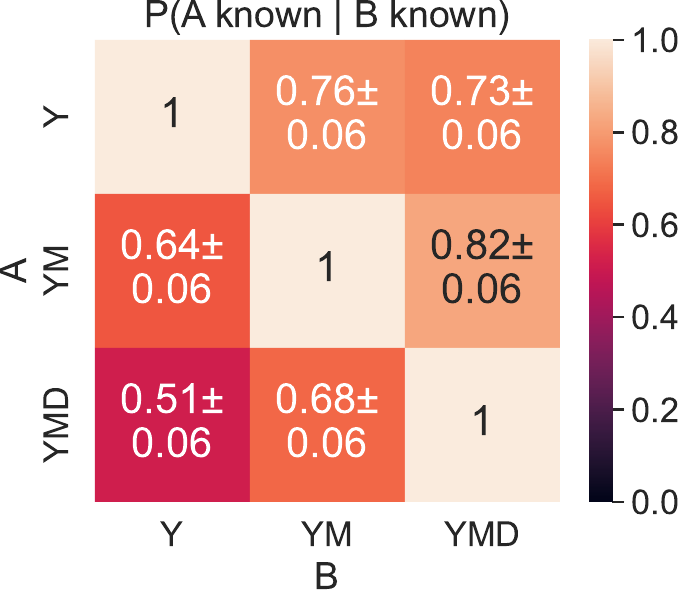}
    \end{subfigure}
    \begin{subfigure}[h]{0.195\linewidth}
        \centering
        \includegraphics[width=\linewidth]{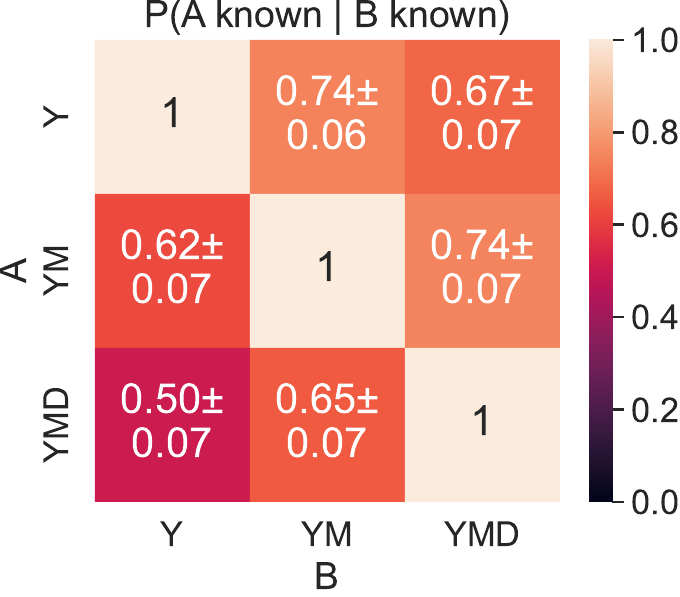}
    \end{subfigure}
    
    \begin{subfigure}[h]{0.195\linewidth}
        \centering
        \includegraphics[width=\linewidth]{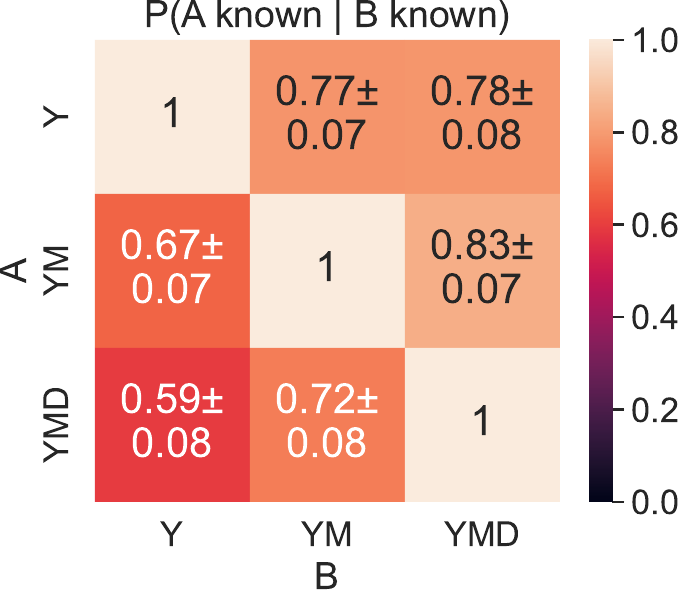}
        \caption{gemma-2-27b-it}
    \end{subfigure}
    \begin{subfigure}[h]{0.195\linewidth}
        \centering
        \includegraphics[width=\linewidth]{plots/gran_vs_smallcomb_noninst_google_gemma-2-9b-it_classic.pdf}
        \caption{gemma-2-9b-it}
    \end{subfigure}
    \begin{subfigure}[h]{0.195\linewidth}
        \centering
        \includegraphics[width=\linewidth]{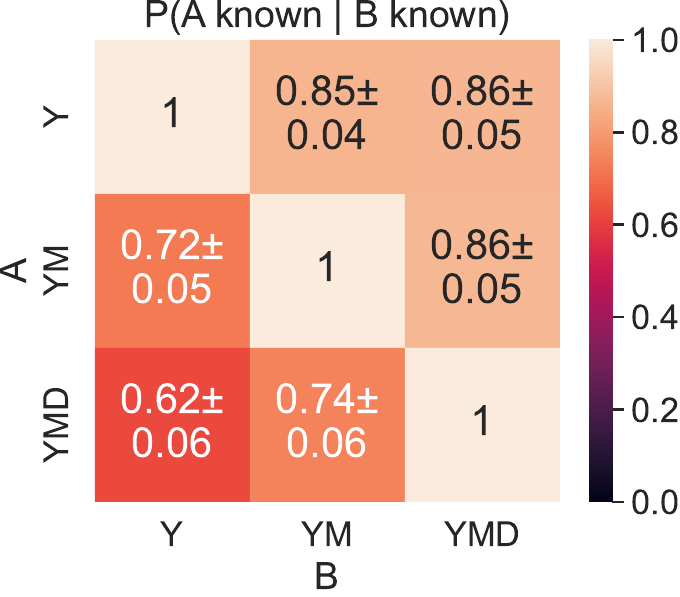}
        \caption{Llama-3.1-70B-Instruct}
    \end{subfigure}
    \begin{subfigure}[h]{0.195\linewidth}
        \centering
        \includegraphics[width=\linewidth]{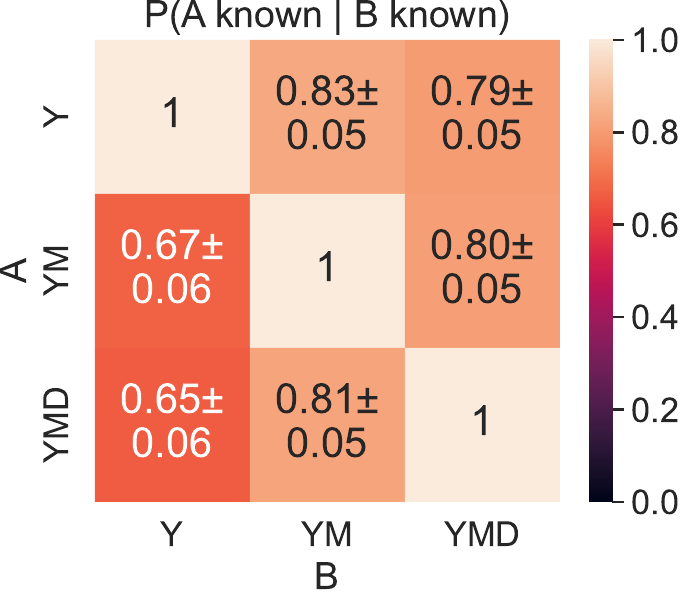}
        \caption{Mistral-Nemo-Instruct-2407}
    \end{subfigure}
    \begin{subfigure}[h]{0.195\linewidth}
        \centering
        \includegraphics[width=\linewidth]{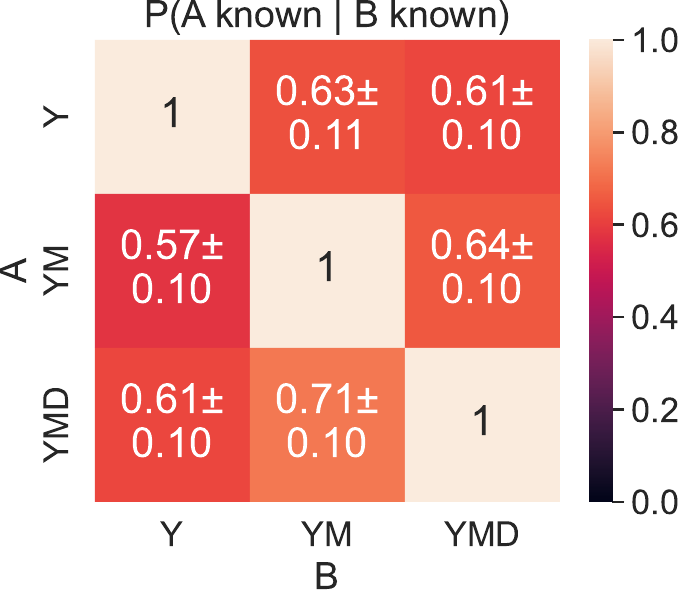}
        \caption{Mistral-7B-Instruct-v0.3}
    \end{subfigure}
    \caption{Generalization matrics between pairs of granularities on the 5 most robust LMs. In the \textbf{first row}, the statements are presented in a \textbf{raw} format, and in the \textbf{second row}, they are presented in a \textbf{instruction} format.}
    \label{fig:gran_gen_all}
\end{figure*}

In Section \ref{sec:anomalies}, we explored the ability of language models to generalize their temporal knowledge from one granularity to another. We provided two matrices (one for instruction-based questions and one for raw text questions) containing the generalization rate between each granularity pair averaged over the 5 most robust LMs. Complementing these average performances, the generalization rate matrices for individual models are presented in Figure \ref{fig:gran_gen_all}.

\subsection{Explanatory Prompts}
\label{appendix:explanation_prompts}
In section \ref{sec:anomalies}, we investigated whether including explanations of temporal concepts in the prompt could help LMs better generalize their knowledge across granularities. Two prompts prefixed to each instruction in TimeStress were used:\\

\noindent\textbf{Prompt 1 : Hierarchical natures of dates}
\small
\begin{tcolorbox}[width=\columnwidth,colback=white]
\tt A date is a specific point in time, expressed through a year, a month, and a day. A year is divided into months, and a month is divided into days. Answer the following question.
\end{tcolorbox}
\normalsize

\noindent\textbf{Prompt 2 : Knowledge transfer between granularities}
\small
\begin{tcolorbox}[width=\columnwidth,colback=white]
\tt A date is a specific point in time. If a fact is valid for a specific year, it holds true for all dates within that year. If a fact is valid for a specific month of a specific year, it holds true for all dates within that month. Answer the following question.
\end{tcolorbox}
\normalsize

The first explains the hierarchical nature of dates, while the second is more straightforward and explains how knowledge of a temporal fact can be generalized across granularities.

\begin{figure*}
    \begin{subfigure}[h]{0.245\linewidth}
        \centering
        \includegraphics[width=\linewidth]{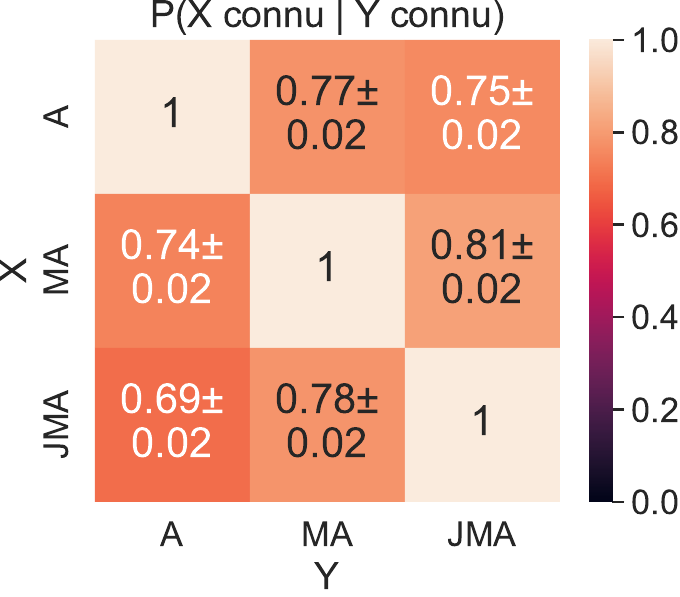}
        \caption{Prompt 1, raw text}
    \end{subfigure}
    \begin{subfigure}[h]{0.245\linewidth}
        \centering
        \includegraphics[width=\linewidth]{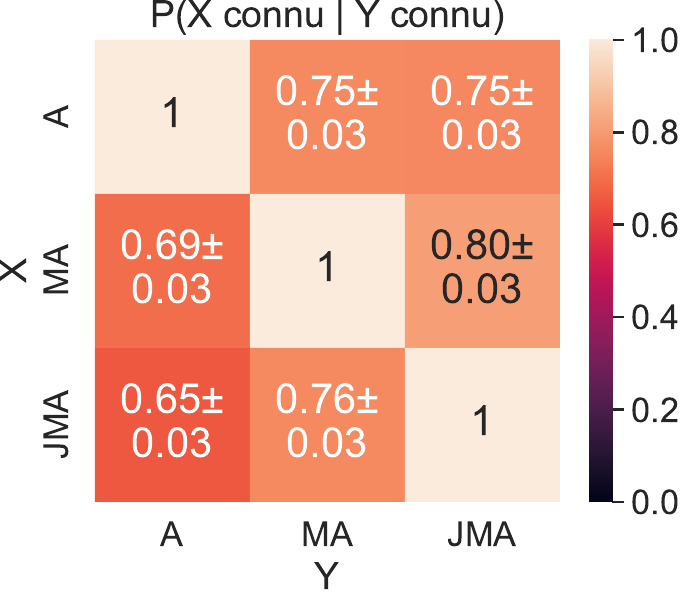}
        \caption{Prompt 1, instruction format}
    \end{subfigure}
    \begin{subfigure}[h]{0.245\linewidth}
        \centering
        \includegraphics[width=\linewidth]{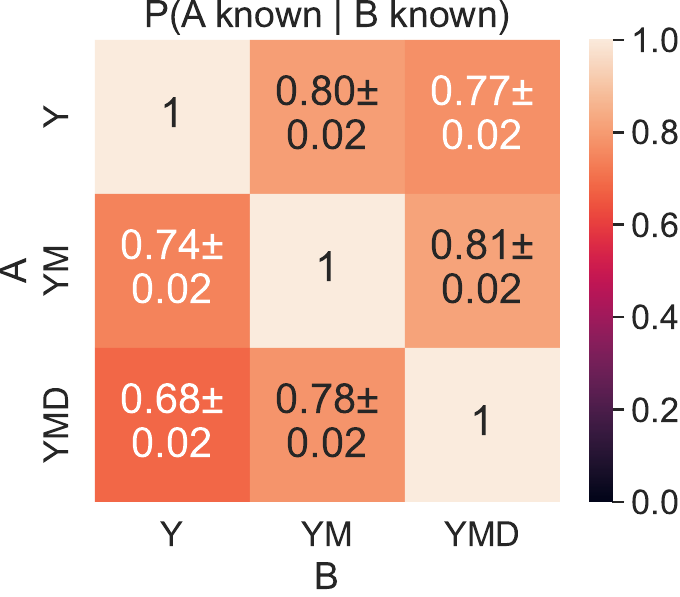}
        \caption{Prompt 2, raw text}
    \end{subfigure}
    \begin{subfigure}[h]{0.245\linewidth}
        \centering
        \includegraphics[width=\linewidth]{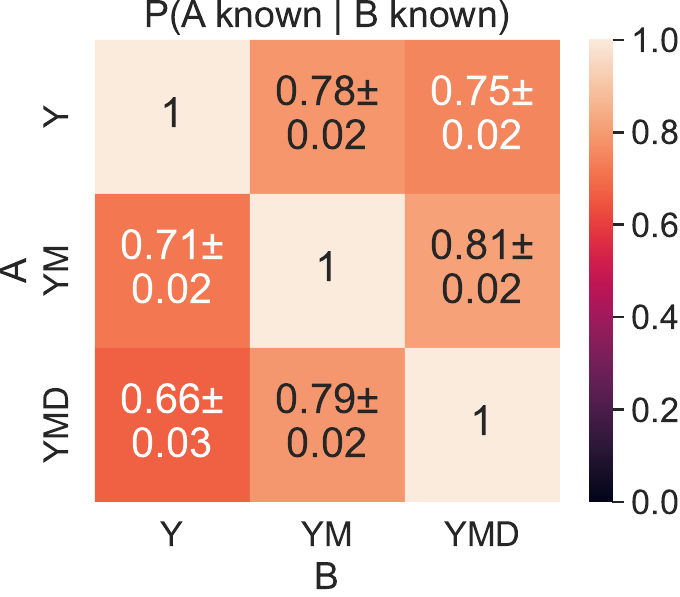}
        \caption{Prompt 2, instruction format}
    \end{subfigure}
    \centering
    \begin{subfigure}[h]{0.2\linewidth}
        \centering
        \includegraphics[width=\linewidth]{plots/gran_vs_smallcomb_noninst_alllms_classic.pdf}
        \caption{No explanation prompt, raw text}
    \end{subfigure}
    \begin{subfigure}[h]{0.245\linewidth}
        \centering
        \includegraphics[width=\linewidth]{plots/gran_vs_smallcomb_inst_alllms_classic.pdf}
        \caption{No explanatory prompt, instruction format}
    \end{subfigure}
    \caption{Effect of adding explanations on temporal concepts through an explanatory prompt}
    \label{fig:gran_gen_explanation_prompts}
    
\end{figure*}

In addition to the average performance in the \ref{sec:anomalies} section, figure \ref{fig:gran_gen_explanation_prompts} shows the average generalization matrices across the same 5 models as in figure \ref{fig:gran_gen}, using raw text and an instruction format.

\section{Conditional Probability Calculations in LMs}
\label{app:prob}
Since our experiments rely entirely on the calculation (by the LM) of the conditional probability of one text given another, it is crucial that these calculations are rigorously implemented.

Given that different tokenizers split a text differently, we require a universal algorithm to best calculate the probability of generating a text given a prompt, even when the end of the prompt might be in the middle of a token.

\noindent Below are the general steps we used to compute $P(A \mid B)$ where $A$ and $B$ are strings:
\begin{enumerate}
\item Tokenize $A+B$ into a sequence of tokens $s=(t_1, t_2, \dots, t_n)$\footnote{$+$ represents the string concatenation operation.}.
\item Find the smallest token sequence $(t_k, \dots, t_n)$ in $s$ that contains $B$, starting from the end.
\item Compute $P(t_k, \dots, t_n \mid t_1, \dots, t_{k-1})$, which can be done using the \textit{logits} produced by the LM.
\end{enumerate}

Other considerations, such as the automatic addition of special tokens by the tokenizer, must also be accounted for. A detailed implementation of this method (the function \texttt{LanguageModel.credibility\_text}) that handles these details is available in the source code.




\begin{figure*}[t]
    \centering
    \begin{subfigure}[h]{0.45\linewidth}
        \includegraphics[width=1\linewidth]{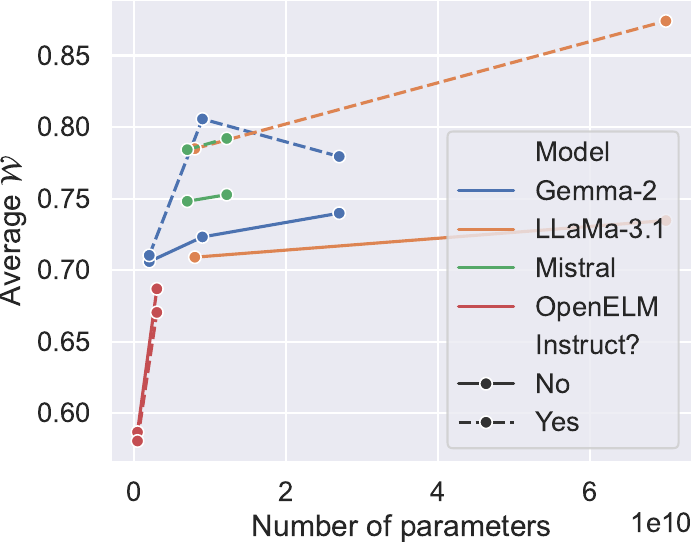}
        \caption{Average Win rate}
        \label{fig:numparams_vs_wr}
    \end{subfigure}\quad\begin{subfigure}[h]{0.45\linewidth}
        \includegraphics[width=1\linewidth]{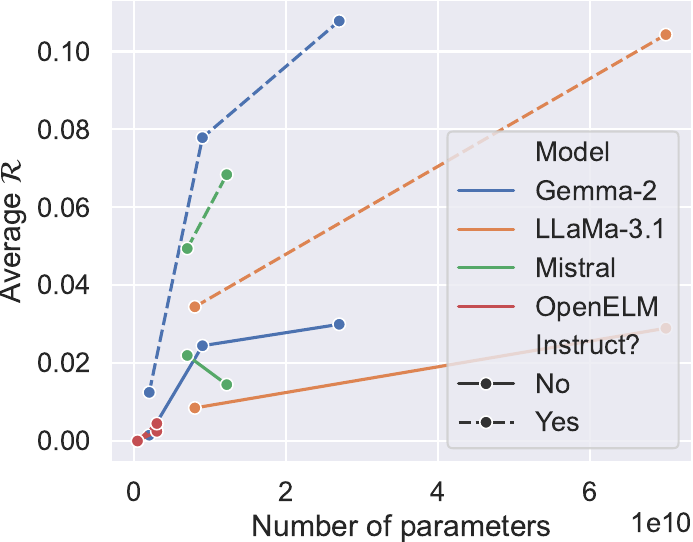}
        \caption{Average Robustness}
        \label{fig:numparams_vs_robust}
    \end{subfigure}
    \caption{Relationship between the number of parameters in an LM and the metric used (across all granularities Y, YM and YMD). Pretrained models are represented by straight lines, while models finetuned on instructions are represented by dotted lines.}
    \label{fig:numparams_vs_perf}
\end{figure*}

\begin{figure*}[t]
    \centering
    \begin{subfigure}[h]{0.48\linewidth}
        \includegraphics[width=1\linewidth]{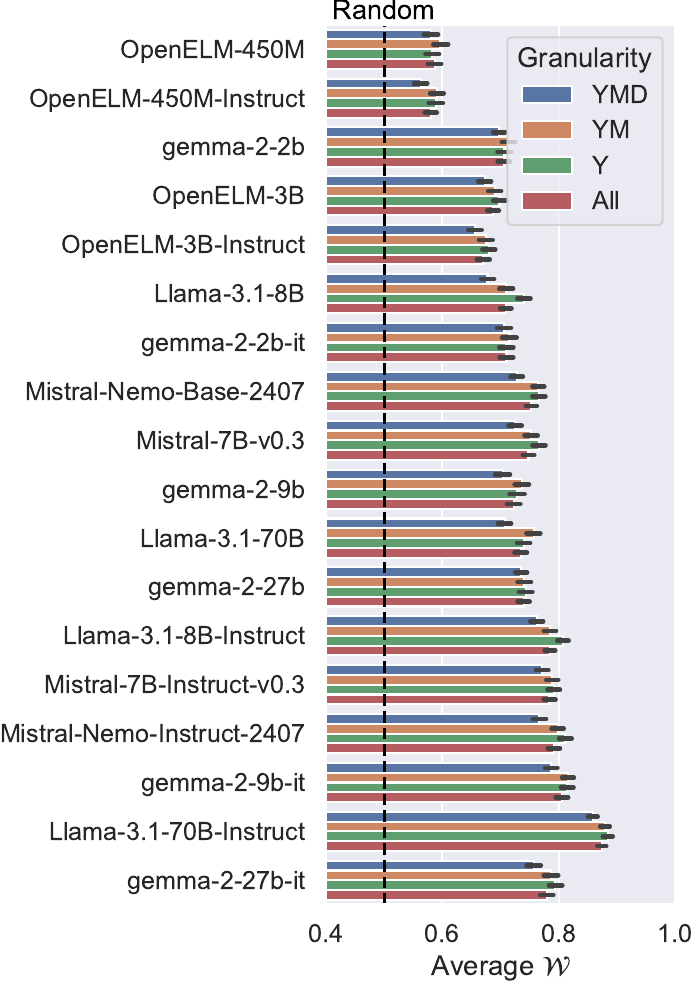}
        \caption{Average win rate}
        \label{fig:full_win_rate}
    \end{subfigure}\quad\begin{subfigure}[h]{0.44\linewidth}
        \includegraphics[width=1\linewidth]{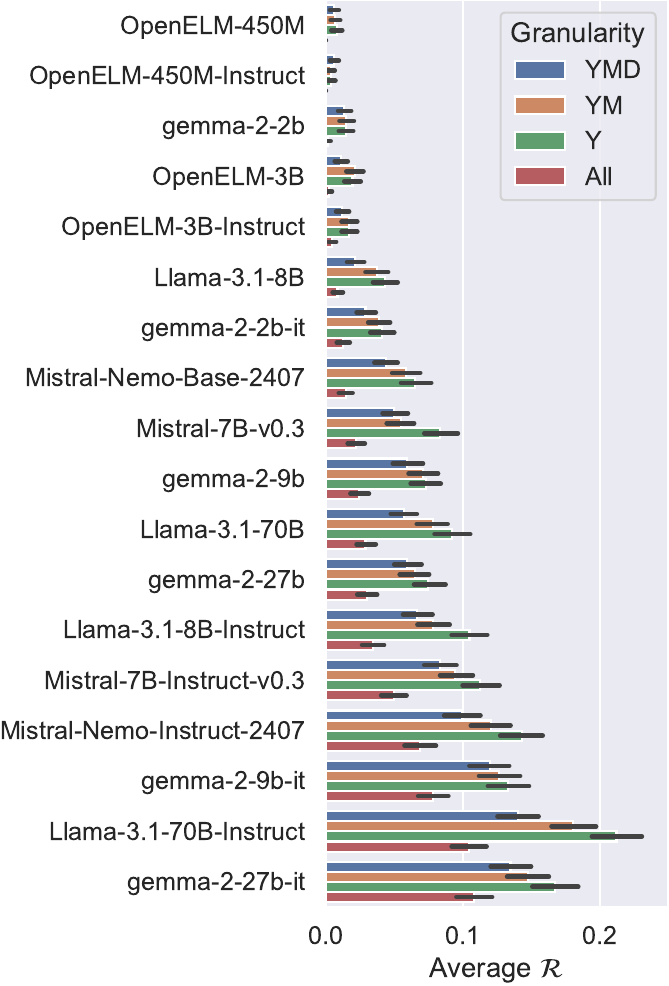}
        \caption{Average robustness}
        \label{fig:full_robust}
    \end{subfigure}
    \caption{Average metrics across all facts in TimeStress for the 18 studied LMs with 95\% confidence intervals (determined using bootstrapping).}
    \label{fig:robust_winrate}
\end{figure*}


\section{Supplementary Results}
\label{appendix:additional_results}

\begin{figure*}[h]
\begin{minipage}{0.45\textwidth}
\includegraphics[width=\linewidth]{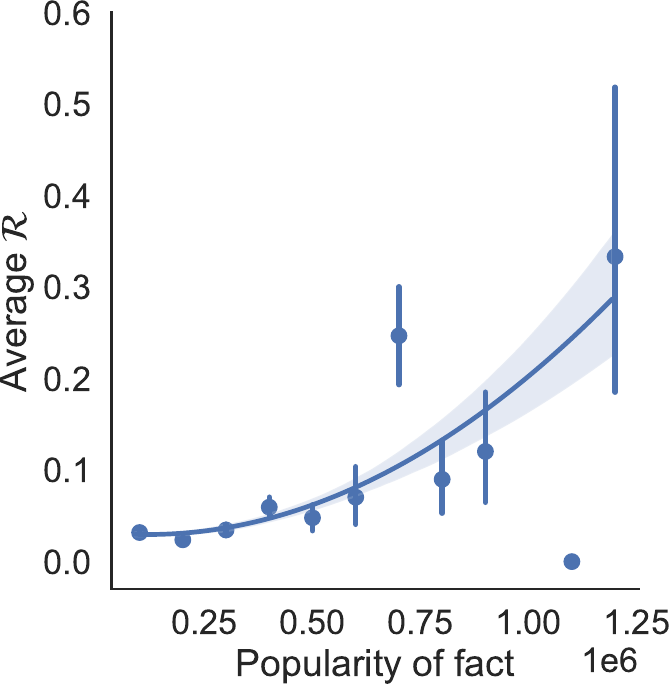}
    \caption{Relationship between fact popularity and robustness metric calculated across granularities Y, YM, YMD. The Pearson coefficient is equal $+0.065$ (p-value $< 10^{-51}$).}
    \label{fig:pop_vs_robust}
\end{minipage}
\hfill
\begin{minipage}{0.48\textwidth}
    \includegraphics[width=\linewidth]{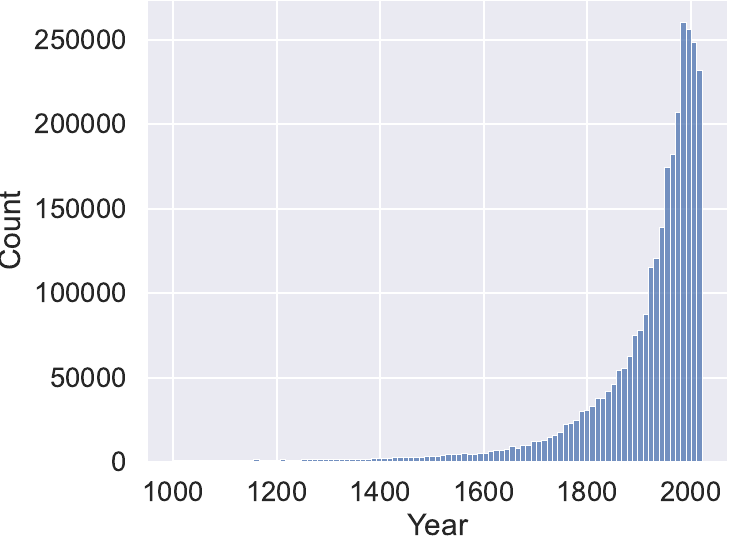}
    \caption{Distribution of the the years of all the temporal contexts in TimeStress.}
    \label{fig:year_dist}
\end{minipage}
\end{figure*}

\begin{figure*}[h]
\begin{minipage}{0.55\textwidth}
    \includegraphics[width=\linewidth]{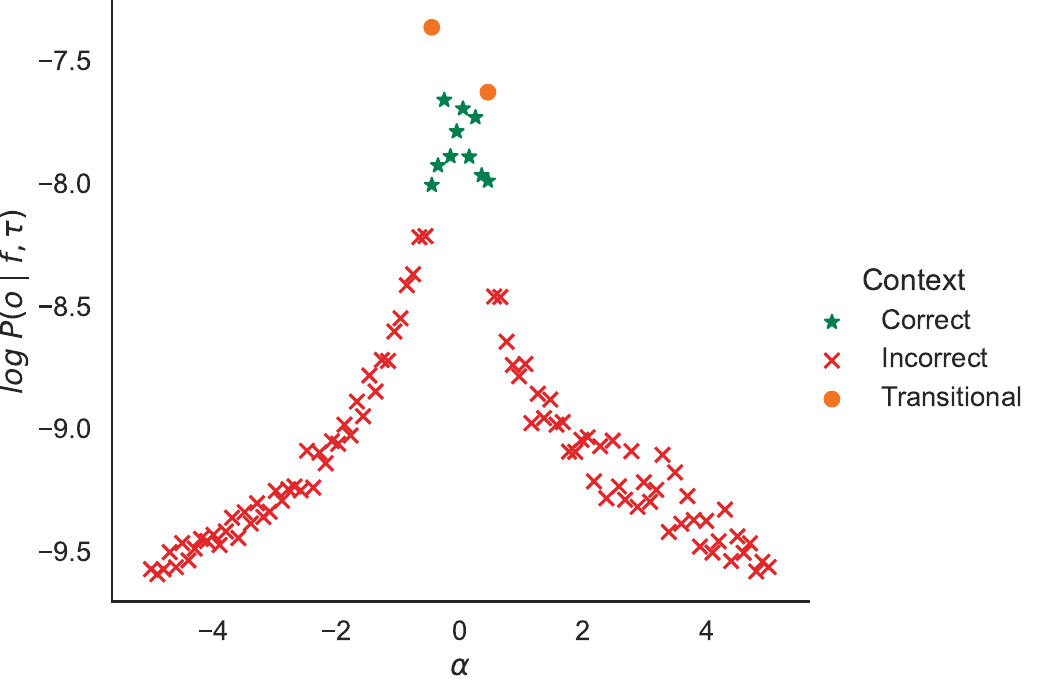}
    \caption{The evolution of $log P (o | f, \tau)$ with respect to the relative distance of the context from the validity period $\alpha$. Each point is an average over many data points.}
    \label{fig:alpha_vs_logprob_detailed}
\end{minipage}
\hfill
\begin{minipage}{0.4\textwidth}
    \includegraphics[width=\linewidth]{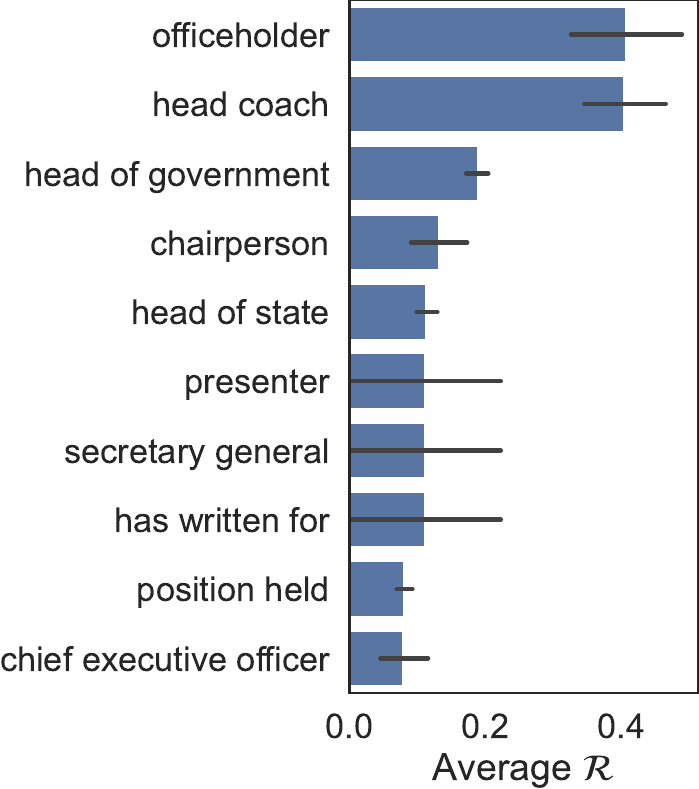}
    \caption{The 10 most known relationships (across all granularities) in TimeStress on average by the studied LMs.}
    \label{fig:top_rels}
\end{minipage}
\end{figure*}

\begin{figure*}
    \centering
    \includegraphics[width=\linewidth]{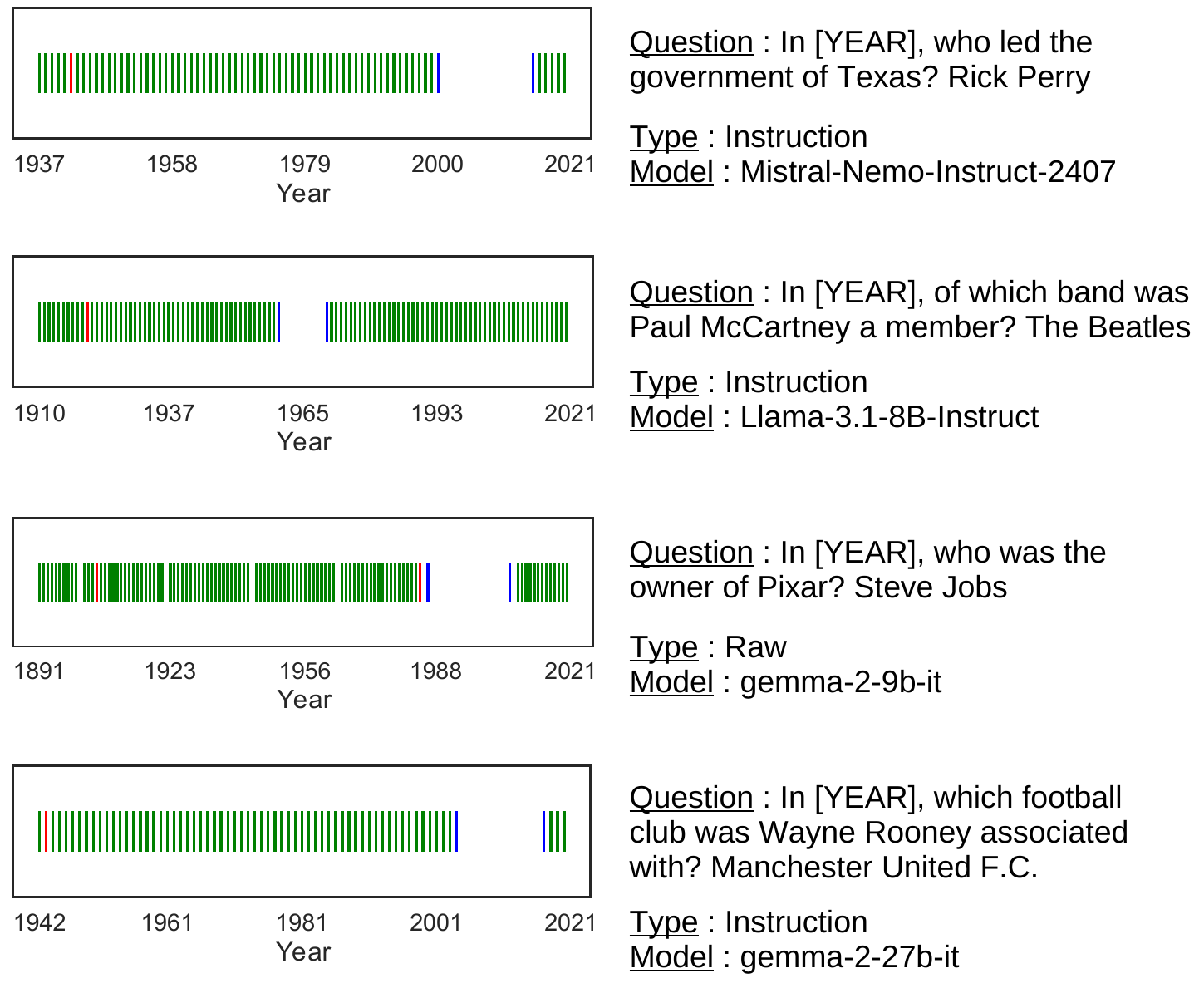}
    \caption{Examples of vulnerability to \textit{easy} incorrect contexts for different LMs. The color {\color{blue}blue} represents the boundaries of the validity period, the color {\color{Green}green} represents incorrect contexts that are never preferred to correct contexts, and the color {\color{red}red}, on the contrary, represents incorrect contexts that were preferred to one or more correct contexts.} 
    \label{fig:example_easy_incorrect_context}
\end{figure*}

\begin{figure*}[t]
    \centering
    \begin{subfigure}[h]{0.45\linewidth}
        \includegraphics[width=1\linewidth]{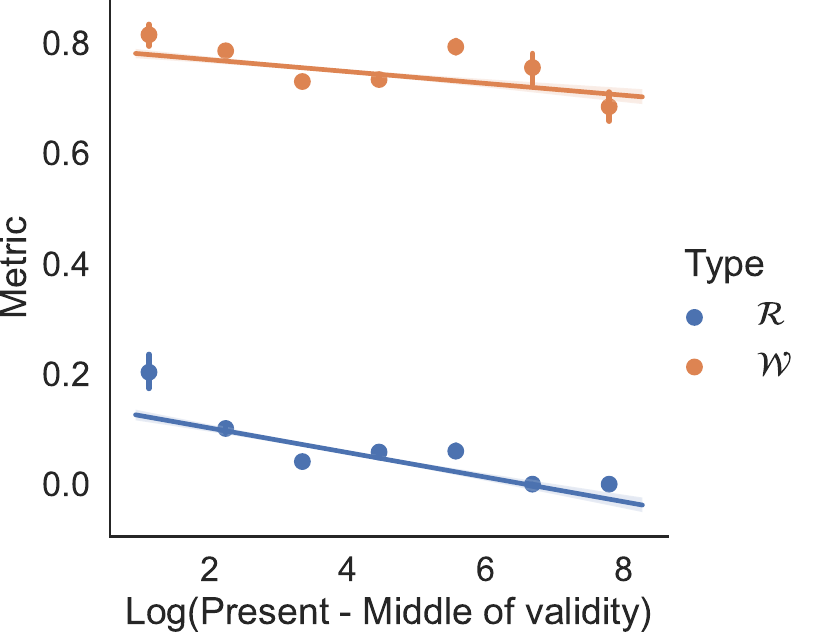}
        \caption{Logarithm of the distance of the fact with respect to the present.}
    \end{subfigure}\quad\begin{subfigure}[h]{0.45\linewidth}
        \includegraphics[width=1\linewidth]{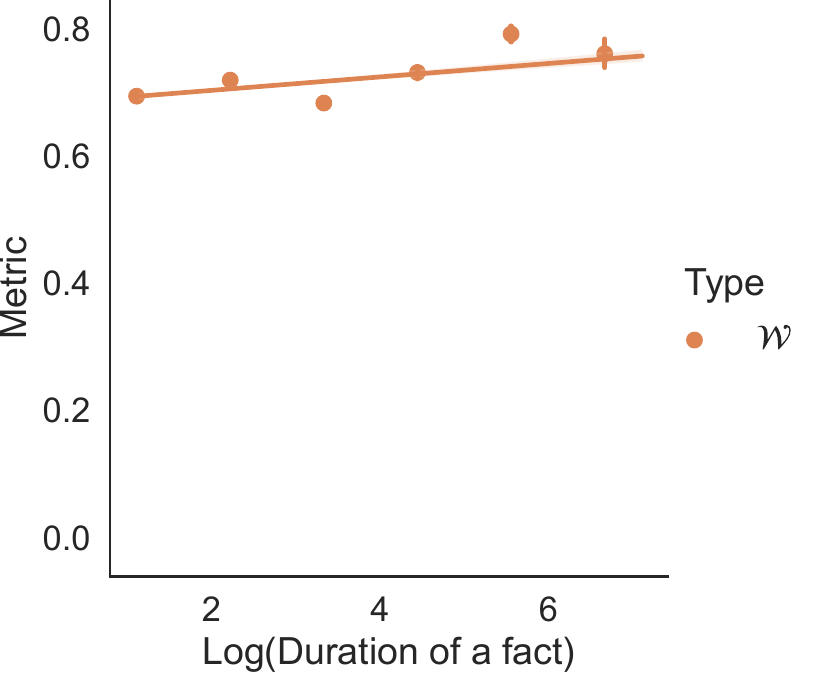}
        \caption{Logarithm of the duration of the fact.}
    \end{subfigure}
    \caption{The influence of two factors on the robustness and win rate of the 5 most robust LMs. All correlations are statistically significant where the null hypothesis is the absence of linear correlation. Robustness is missing from Figure b because its analysis is not relevant as the duration of a fact is confounded with another variable: the number of matches of a fact. Indeed, the longer a fact is, the more matches it has, and the lower is the robustness.}
    \label{fig:dist_duration_vs_metrics}
\end{figure*}

\begin{itemize}
\item The average robustness score and win rate across the 18 studied LMs are presented in Figure \ref{fig:robust_winrate}.

\item The relationship between the number of parameters in LMs and their performance is shown in Figure \ref{fig:numparams_vs_perf}.

\item Figure \ref{fig:alpha_vs_logprob_detailed} illustrates the evolution of $log P(o \mid f, \tau)$ with respect to the relative distance of the date from the validity period $\alpha$, which is equivalent to Figure \ref{fig:alpha_vs_logprob} but with more details.

\item Figure \ref{fig:top_rels} displays the relations that were most robustly known on average by the studied LMs ("raw text" format statements).


\item Figure \ref{fig:example_easy_incorrect_context} shows examples where LMs were vulnerable to easy incorrect contexts.

\item Figure \ref{fig:year_dist} shows the year distribution of temporal contexts across the entire TimeStress dataset.

\item Figure \ref{fig:dist_duration_vs_metrics} shows the influence of fact distance from the present (here, the year 2021), as well as their durations, on the robustness and win rate of the 5 most robust MLs. The time unit used for both metrics is the \textbf{year}.

\end{itemize}

\begin{table*}[h]
\resizebox{\linewidth}{!}{

\begin{tabular}{@{}p{10cm}p{8cm}l@{}}
\toprule
\textbf{Temporal Fact} & \textbf{Statement} & \textbf{Status} \\
\midrule
(Alexander Graham Bell, country of citizenship, United States of America, 1882, 1922) & In July 1734, what was Alexander Graham Bell's country of citizenship\string? United States of America & Incorrect \\
\midrule
(Lauren Bacall, spouse, Jason Robards, 1961-07-04, 1969-09-10) & In July 1984, who was the spouse of Lauren Bacall\string? Jason Robards & Incorrect \\
\midrule
(Vatican City, head of state, John Paul II, 1978-10-16, 2005-04-02) & In July 2006, who held the highest authority in Vatican City\string? John Paul II & Incorrect \\
\midrule
(Gareth Barry, member of sports team, Manchester City F.C., 2009, 2014) & In July 2020, which football team included Gareth Barry as a player\string? Manchester City F.C. & Incorrect \\
\midrule
(Pierce Brosnan, spouse, Cassandra Harris, 1980, 1991) & In 1954, who did Pierce Brosnan have as his wife\string? Cassandra Harris & Incorrect \\
\midrule
(Metallica, has part, Jason Newsted, 1987, 2001-01-17) & In 1971, who was included in Metallica's lineup\string? Jason Newsted & Incorrect \\
\midrule
(Eliza Dushku, unmarried partner, Rick Fox, 2009, 2014) & In 2003, who was Eliza Dushku in a relationship with\string? Rick Fox & Incorrect \\
\midrule
(United Kingdom, head of state, George VI, 1936-12-11, 1952-02-06) & On July 1, 1892, who served as the king of the United Kingdom\string? George VI & Incorrect \\
\midrule
(Linda Lee Cadwell, spouse, Bruce Lee, 1964, 1973-07-20) & In 1929, who was the spouse of Linda Lee Cadwell\string? Bruce Lee & Incorrect \\
\midrule
(George Harrison, part of, The Beatles, 1960, 1970) & On July 2, 1971, what was the name of the band that George Harrison was associated with\string? The Beatles & Incorrect \\
\midrule
(Philippines, head of state, Corazon Aquino, 1986-02-25, 1992-06-30) & On July 2, 1969, who served as the leader of the Philippines\string? Corazon Aquino & Incorrect \\
\midrule
(Jawaharlal Nehru, position held, Prime minister of India, 1947-08-15, 1964-05-27) & In 1985, what position did Jawaharlal Nehru hold\string? Prime Minister of India & Incorrect \\
\midrule
(Vienna, country, Austria-Hungary, 1867-03-30, 1918-11-11) & In July 1769, which country did Vienna belong to\string? Austria-Hungary & Incorrect \\
\midrule
(Mileva Marić, spouse, Albert Einstein, 1903, 1919) & In July 1907, who was Mileva Marić married to\string? Albert Einstein & Correct \\
\midrule
(Mayte Garcia, spouse, Prince, 1996, 2000) & In July 1979, who was the spouse of Mayte Garcia\string? Prince & Incorrect \\
\midrule
(Abkhazia, country, Soviet Union, 1921, 1991) & In July 1956, which country did Abkhazia belong to\string? Soviet Union & Correct \\
\midrule
(Georgia, member of, Commonwealth of Independent States, 1993-12-03, 2009-08-18) & In 1930, what group included Georgia as a member\string? Commonwealth of Independent States & Incorrect \\
\midrule
(Abraham Lincoln, member of political party, Whig Party, 1834, 1854) & In 1808, which political party was Abraham Lincoln a member of\string? Whig Party & Incorrect \\
\midrule
(Wales, located in the administrative territorial entity, Kingdom of England, 1284, 1707-04-30) & On July 1, 1072, which territorial entity included Wales\string? Kingdom of England & Incorrect \\
\midrule
(Frédéric Chopin, residence, Paris, 1831, 1849) & On July 2, 1847, which city was home to Frédéric Chopin\string? Paris & Correct \\
\bottomrule
\end{tabular}

}
\caption{Random sample from TimeStress.}\label{tab:timestress_sample}
\end{table*}

\end{document}